\documentclass[journal]{IEEEtran}

\usepackage{algorithm}
\usepackage{algorithmic}
\usepackage{mathtools}
\usepackage{amsfonts}
\usepackage{amssymb}
\usepackage{subfig}
\usepackage{rotating}
\usepackage[table]{xcolor}

\definecolor{light-gray}{gray}{0.9}

\ifCLASSINFOpdf
\else
\fi
\begin{document}
\title{WoCE: a framework for clustering ensemble by exploiting the wisdom of Crowds theory}

\author{Muhammad~Yousefnezhad,
	Sheng-Jun Huang,
        Daoqiang~Zhang
\thanks{M. Yousefnezhad is with the College
of Computer Science and Technology, Nanjing University of Aeronautics and Astronautics,  Nanjing,
Jiangsu, 210016 China e-mail: (myousefnezhad@nuaa.edu.cn).}% <-this % stops a space
\thanks{S.~J. Huang is with the College
	of Computer Science and Technology, Nanjing University of Aeronautics and Astronautics,  Nanjing,
	Jiangsu, 210016 China e-mail: (huangsj@nuaa.edu.cn).}% <-this % stops a space
\thanks{D. Zhang is with the College
	of Computer Science and Technology, Nanjing University of Aeronautics and Astronautics,  Nanjing,
	Jiangsu, 210016 China e-mail: (dqzhang@nuaa.edu.cn).}% <-this % stops a space
}

%\author{Michael~Shell,~\IEEEmembership{Member,~IEEE,}
%	John~Doe,~\IEEEmembership{Fellow,~OSA,}
%	and~Jane~Doe,~\IEEEmembership{Life~Fellow,~IEEE}% <-this % stops a space
%	\thanks{M. Shell is with the Department
%		of Electrical and Computer Engineering, Georgia Institute of Technology, Atlanta,
%		GA, 30332 USA e-mail: (see http://www.michaelshell.org/contact.html).}% <-this % stops a space
%	\thanks{J. Doe and J. Doe are with Anonymous University.}% <-this % stops a space
%	\thanks{Manuscript received April 19, 2005; revised September 17, 2014.}}



\markboth{IEEE Transaction on Cybernetics}%
{Shell \MakeLowercase{\textit{et al.}}: Bare Demo of IEEEtran.cls for Journals}




\maketitle

\begin{abstract}
The Wisdom of Crowds (WOC), as a theory in the social science, gets a new paradigm in computer science. The WOC theory explains that the aggregate decision made by a group is often better than those of its individual members if specific conditions are satisfied. This paper presents a novel framework for unsupervised and semi-supervised cluster ensemble by exploiting the WOC theory. We employ four conditions in the WOC theory, i.e., diversity, independency, decentralization and aggregation, to guide both the constructing of individual clustering results and the final combination for clustering ensemble. Firstly, independency criterion, as a novel mapping system on the raw data set, removes the correlation between features on our proposed method. Then, decentralization as a novel mechanism generates high quality individual clustering results. Next, uniformity as a new diversity metric evaluates the generated clustering results. Further, weighted evidence accumulation clustering method is proposed for the final aggregation without using thresholding procedure. Experimental study on varied data sets demonstrates that the proposed approach achieves superior performance to state-of-the-art methods.
\end{abstract}


% \fi
\IEEEpeerreviewmaketitle

% \IEEEPARstart{\top}{his demo} file is ....
%\IEEEPARstart{T}{his} demo file is intended to serve as a ``starter file''
%for IEEE journal papers produced under \LaTeX\ using
%IEEEtran.cls version 1.8a and later.
%% You must have at least 2 lines in the paragraph with the drop letter
%% (should never be an issue)
%I wish you the best of success.

\section{Introduction}
\IEEEPARstart{C}{lustering}, the art of discovering meaningful patterns in the unlabeled data sets, is one of the main tasks in machine learning. Semi-supervised clustering is a branch of clustering methods that uses prior supervision information, such as labeled data, known data associations, or pairwise constraints, to aid the clustering process. This paper focuses on pairwise constraints, i.e. pairs of instances known as belonging to the same cluster (must-link constraints) or different clusters (cannot-link constraints). Pairwise constraints arise naturally in many real tasks and have been widely used in semi-supervised clustering. There is a wide range of issues in the clustering methods. For instance, individual clustering algorithms provide different accuracies in a complex data set because they generate the clustering results by optimizing a local or global function instead of natural relations between data points \cite{Strehl02, Topchy03,Fred08,Jain04}. As another example, pairwise constraints often result in highly unstable clustering performance, whereas they have the potential to improve clustering accuracy in practice \cite{Anand14,Xiong14}. 

As a novel solution, cluster ensemble was proposed for achieving a robust and stable final result by combining the different individual clustering results \cite{Strehl02}. Cluster Ensemble Selection (CES) is a new approach which combines a subgroup of individual clustering results. It uses a consensus metric(s) for evaluating and selecting the ensemble committee in order to improve the accuracy of final results \cite{Fern08}. Generally, CES contains four components; i.e. generation, evaluation, selection, and combination. Firstly, individual clustering results are generated by using different kinds of clustering algorithms or repeating some algorithms, which can generate random results in each runtime such as the k-means. Next, a consensus metric(s) such as Normalized Mutual Information (NMI) is employed to evaluate the generated results. After that, the evaluated results are selected by thresholding procedure. Lastly, the final clustering result is obtained by an aggregation mechanism \cite{Fern08,Azimi09,Alizadeh14,Limin12, Jia12}. 

There are three challenges in the CES arena; i.e. strategy of generation, metric(s) of evaluation, thresholding procedure. As the first challenge, the strategy of generating the individual clustering results can dramatically affect the performance of CES \cite{Yousefnezhad15,Yu15,Yu14,Huang15,Jing15}. There are generally two paradigms, i.e. some of these studies \cite{Fern08,Alizadeh12,Alizadeh14,Yu15} separately run each component of the CES (generate all individual results, then evaluate them, etc.) whereas the rest of studies \cite{Alizadeh15,Yousefnezhad15} employed \emph{feedback mechanism}, which gradually runs each component of the CES (generating the first individual result, then evaluating it, etc.). On the one hand, feedback mechanism uses evaluated the results at each step for improving the quality of the generated results in the next steps. Therefore, it can usually provide better performance in comparison with the first paradigm \cite{Alizadeh15,Yousefnezhad15}. On the other hand, it may not be compatible with many of classical structures/metrics in the ensemble learning. Evaluation is the next challenge. NMI is one of the most prevalent diversity metrics that is used in the CES because 1) NMI is not sensitive to the cluster's indices \cite{Alizadeh15} 2) it can be easily implemented \cite{Azimi09,Fern08} 3) it has better time complexity in comparison with other classic methods \cite{Fred05,Fern08,Alizadeh12,Alizadeh14}. The main disadvantage of NMI is that the symmetric problem. Indeed, it cannot provide an efficient evaluation while the numbers of instances in distinct clusters are highly different. For instance, consider a clustering analysis for partitioning emails to normal or spam groups, where the number of instances in the normal group is significantly greater than the number of data points in the spams group. Alizadeh et al. \cite{Alizadeh14,Alizadeh12,Alizadeh15} proved that the NMI evaluates the similarity between these two clusters equal to 1, while the real similarity is near to zero. This issue can rapidly decrease the performance of the NMI-based CES methods in the big data analysis \cite{Alizadeh12,Alizadeh14,Alizadeh15,Yousefnezhad15}. Recently, some of the studies proposed a modified version of the NMI such as APMM\footnote{Alizadeh-Parvin-Moshki-Minaei} \cite{Alizadeh14} and MAX \cite{Alizadeh12} for solving this problem. Their proposed methods were utilized for evaluating diversity between a cluster and a partition. Since using mentioned methods for evaluating two partitions increases the time complexity, it is critical to propose a new metric, which directly can evaluate diversity between two partitions. The next challenge in the CES is thresholding. In practice, it is so hard to find optimum values of thresholds; and the performance of the CES significantly depends on the threshold values \cite{Yousefnezhad15}.

Most of the ensemble methods (especially in the CES) employs the (majority) voting systems \cite{Fern08,Azimi09,Alizadeh15,Yousefnezhad15}, such as Boosting and Error-Correcting Output Codes (ECOC) in supervised learning \cite{Tan05} or Evidence Accumulation Clustering (EAC) method in unsupervised learning \cite{Fred05}. Indeed, CES framework just provides a voting system for selecting the robust and stable individual results. Voting systems are firstly defined in the term of social science, where it is used for providing democratic societies, fair trials (in the courts), etc. \cite{Surowiecki04}. There is a wide range of theories in social science, which can provide an environment for applying an effective voting system. They can be used to inspire new algorithms in machine learning. The Wisdom of Crowds (WOC) is one of these theories, which explain a robust approach for generating accurate results in a voting system. It simply claims that decisions made by aggregating the information of groups are better than those made by any single group member if the four specific conditions of this theory are satisfied; i.e. diversity, independency, decentralization, and aggregation \cite{Surowiecki04,Alizadeh15}. Indeed, we can find many modern concepts in different sciences, which used WOC as a fundamental resource, e.g. Delphi method in management \cite{Surowiecki04}, crowdsourcing/funding in the market \cite{Surowiecki04}, crowd computing \cite{Xue12} in computer networks, etc. In computer science, this theory was used for optimizing resources in wireless sensor networks \cite{Xue12}. Further, there is a wide range of studies in supervised learning \cite{Baker08,Miller09,Steyvers09,Welinder10,Williams10,Yi10} and unsupervised learning \cite{Yousefnezhad15,Alizadeh15}, which use the WOC theory for proposing new approaches. These studies validated that the WOC theory usually leads to better \emph{performance} and higher \emph{stability}. 

For solving the three mentioned problems in CES, this paper shows that the WOC theory well matches the target of cluster ensemble, and thus its four conditions can be employed to guide the designing of individual clusterings as well as the final ensemble. Based on this observation, we propose a robust framework, which is called Wisdom of crowds Cluster Ensemble (WoCE), for both unsupervised and \emph{semi-supervised} cluster ensemble. Our contribution in this paper can be summarized as follows: 
\begin{itemize}
\item {Firstly, a new mapping between the WOC observations and the CES problems. Furthermore, a general framework is proposed based on WOC theory for generating diverse individual results and using the feedback mechanism to select individual clusterings with high independency and quality. This framework is the first WOC-based approach for semi-supervised clustering.}
\item {After that, this paper introduces a novel technique in the term of mathematical independent random variables for mapping the data to new dimensions based on the natural correlation of raw data, which can satisfy the \emph{independency} criterion in the WOC. This mapping can generate independent features, which increase the performance of individual clustering algorithms.}
\item{Then, to satisfy the \emph{decentralization} criterion in WOC, this study uses different numbers of clusters in the different kinds of clustering algorithms, which can effectively generate high quality individual clustering results. Moreover, this paper develops a new method for selecting features based on supervision information in the semi-supervised approach.}	
\item{Next, to satisfy the \emph{diversity} criterion in the WOC, this study proposes a new diversity metric called uniformity, which is based on the APMM criterion, for evaluating the diversity of two partition, directly \cite{Alizadeh14}}. 
\item{Lastly, to satisfy the \emph{aggregation} mechanism in WOC, this paper proposed Weighted Evidence Accumulation Clustering (WEAC) to obtain the final clustering with a weighted combination of all individual results. While the weight of each individual result in WEAC can be estimated with different metrics, the uniformity was used in this paper.}
\end{itemize}

The rest of this paper is organized as follows: In Section II, this study first briefly reviews some related works. Then it introduces the proposed WoCE framework in Section III. Experimental results are presented in Section IV; and finally this paper presents conclusion and point out some future works in Section V.
\section{Related Works}
\subsection{The Wisdom of Crowds}
Francis Galton was a British scientist, who introduced the correlation concept in statistics. In 1906, he went to annual West of England Fat Stock and Poultry Exhibition where the local farmers and townspeople gathered to estimate and gamble the quality of each other's cattle, sheep, pigs, etc.  Each animal was shown to the crowd; and people wrote their estimations on the tickets. Final goal of this gambling was estimating the closest weight for each animal in comparison with the real weight of that animal. Galton considered that the average of tickets' value for each animal must be a value of significant distance in comparison with the exact answer because a few people (local farmers or experts) just knew the right answer. He borrowed all 787 tickets, which show the estimations of an ox's weight. While the weight of that ox was 1198 pounds, the average of estimated values in the tickets was 1197! In 1907, he published the `Vox Populi' paper in the Nature journal; and mentioned that “the result seems more creditable to the trustworthiness of a democratic judgment that might have been expected”. In fact, he understood that each ticket contains two data; i.e. information and error. Errors in the tickets omit each other, and the information summarized. This is the main reason that the average of those tickets was really quiet in comparison with the correct answer. This is the core idea of the wisdom of crowds theory in social science. Further, this theory is comparable with the jury theorem, which was proposed by Condorcet. Supported by a wide range of examples in business, management, economic, social science, mathematician, etc., Surowiecki introduced the wisdom of crowds as a framework for making optimized decisions in 2004. He proposed four criteria for a wise crowd: \cite{Surowiecki04}\\
\emph{\textbf{Independency}} People's opinions are not determined by the opinions of those around them.\\
\emph{\textbf{Decentralization}} People are able to specialize and draw on local knowledge.\\
\emph{\textbf{Diversity}} Each person has private information, even if it is only an eccentric interpretation of the known facts.\\
\emph{\textbf{Aggregation}} Some mechanism exists for turning private judgments into a collective decision.

There are some examples for unwise crowds in Surowiecki's book; i.e. Columbia shuttle disaster, bubble in the stock markets, etc. Further, he mentioned to three failures in the crowd intelligence. In other words, the wisdom of crowds cannot solve these types of problems. The first is called ants circular mill, which was introduced by William Beebe. An ant mill is an observed phenomenon in which a group of army ants separated from the main foraging party loses the pheromone track and begins to follow one another, forming a continuously rotating circle. Next is called Needle in a Haystack. In this type of problem, just a few members of a group know the right answer. The last is called random decisions. In this type of problem, the final result is completely generated independent of members' decisions. Although the wisdom of crowds cannot solve the mentioned problems, it is employed in the different fields of science as a novel theory for solving problems. For instance, it is one of the main references for the Delphi method in management, crowd sourcing and funding in business, the problem solving theorem and the central limit theorem in the mathematician, etc. \cite{Surowiecki04}.
\subsection{Cluster Ensemble}
Clustering groups data points into clusters so that members of the same cluster are more similar to each other than to members of other clusters. Semi-supervised clustering uses supervision information to aid the clustering process. This paper focuses on pairwise constraints-based semi-supervised methods. As constraint-based methods: Liu et al. proposed semi-supervised linear discriminant clustering (Semi-LDC) \cite{Liu14}. Wang et al. introduced a new technique by utilizing the constrained pairwise data points and their neighbors, which is denoted as constraint neighborhood projections that required fewer labeled data points (constraints) and can naturally deal with constraint conflicts \cite{Wang14}. Chen et al. recently proposed a clustering algorithm which is based on graph clustering and optimizing an appropriately weighted objective, where larger weights are given to observations with lower uncertainty \cite{Chen14}. 

As mentioned before, individual clustering algorithms provide predictions with different accuracy rates. In practice, individual algorithms may fail to provide accurate and stable results. For solving this problem, cluster ensemble proved that better final results can be generated by combining individual clustering results instead of only choosing the best one \cite{Strehl02}. The idea that not all partitions are suitable for cooperating to generate the final clustering was proposed in Cluster Ensemble Selection (CES). This method combines a selected group of best individual clustering results according to consensus metric(s) from the ensemble committee in order to improve the accuracy of final results \cite{Fern08}. 

There are a wide range of studies in the unsupervised cluster ensemble (selection). Vega et al. proposed Weighted Partition Consensus via Kernels (WPCK) method, which analyzes the set of partitions in the cluster ensemble and extracts valuable information that can improve the quality of the combination process \cite{Vega10}. In another study, Vega et al. developed the Weighted Evidence Accumulation (WEA) algorithm by computing the weighted association matrix as the first step and after that, applying a hierarchical clustering algorithm for selecting the consensus partition with the highest lifetime criterion. They also introduced the Generalized Kernel Partition Consensus (GKPC) method that uses the Information Unification step after the generation in the methodology of the WPCK method \cite{Vega11}. Jia et al. proposed SIM for diversity measurement, which works based on the NMI \cite{Jia12}. Romano et al. proposed Standardized Mutual Information (SMI) for evaluating clustering results \cite{Romano14}. Yu et al. proposed the Hybrid Clustering Solution Selection (HCSS) strategy that utilizes a weighting function to combine several feature selection techniques for the refinement of clustering solutions in the ensemble \cite{Yu14}. Based on Normalized Crowd Agreement Index (NCAI) and multi-granularity information collected among individual clusterings, clusters, and data instances, Huang et al. proposes two novel consensus functions, termed weighted evidence accumulation clustering (WEAC) and graph partitioning with multi-granularity link analysis (GP-MGLA) \cite{Huang14}. Jing et al. introduced a component generation approach for producing ensemble components based on Stratified feature sampling \cite{Jing15}. Yu et al. adopted affinity propagation (AP) in different subspaces of the data set for generating a set of individual clusterings \cite{Yu15}.    Alizadeh et al. have concluded the disadvantages of NMI as a symmetric criterion. They used the APMM and Maximum (MAX) metrics to measure diversity and stability, respectively, and suggested a new method for building a co-association matrix from a subset of the individual cluster results. While the proposed methods can solve the symmetric problem of the NMI method, they just can combine a sub-clusters of the generated partition in the reference set \cite{Alizadeh12, Alizadeh14}. Yousefnezhad et al. proposed Weighted Spectral Cluster Ensemble (WSCE) method by exploiting the concept of community detection and graph based clustering \cite{Yousefnezhad15}. 

Gao et al. introduced a graph-based consensus maximization (BGCM) method for combining multiple supervised and unsupervised models. This method consolidated a classification solution by maximizing the consensus among both supervised predictions and unsupervised constraints. Since, this research used a classification approach for unsupervised learning, it is sensitive to the quality of supervision information \cite{Gao13}. Huang et al. extended extreme learning machines (ELMs) for both semi-supervised and unsupervised tasks based on the manifold regularization \cite{Huang14}. Anand et al. proposed a semi-supervised framework for kernel mean shift clustering (SKMS) that uses only pairwise constraints to guide the clustering procedure. They used the initial kernel matrix by minimizing a LogDet divergence-based objective function for first mapped to a high-dimensional kernel space where the constraints are imposed by a linear transformation of the mapped points \cite{Anand14}. Xiong et al. proposed Neighborhood-based Framework (NBF) method. This method builds on the concept of neighborhood, where neighborhoods contain “labeled examples” of different clusters according to the pairwise constraints. Furthermore, it expands the neighborhoods by selecting informative points and querying their relationship with the neighborhoods \cite{Xiong14}.

One of the biggest challenges in the mentioned methods is that they did not use the achieved errors, i.e. false positive and false negative, for improving the quality of the final aggregation. As mentioned before, WOC theory uses information and errors for increasing the performance of the final result. Briefly, information aggregate with each other; and also, errors omit each other. There are several studies based on the WOC theory in supervised learning, e.g. in recollecting ordering information \cite{Steyvers09}, rank ordering problem \cite{Miller09}, estimating the underlying value (e.g., the class) in the image processing \cite{Welinder10}, underwater mine classification with imperfect labels \cite{Williams10}, minimum spanning tree problems \cite{Yi10}, and classification ensemble  \cite{Baker08}. As the first WOC-based unsupervised CES method, Alizadeh et al. proposed the Wisdom of Crowds Cluster Ensemble (WOCCE) . They proposed a new strategy of generating, evaluating, selecting, and combining the individual clustering results based on WOC theory. The main advantages of the WOCCE are using feedback mechanism for managing errors in each iteration and utilizing the A3 metric (average of APMM) to avoid the NMI symmetric problem. There are also four disadvantages in the WOCCE method. Firstly, WOCCE needs three distinct kinds of threshold values for generating final clustering result. Further, the performance of WOCCE is dramatically sensitive to the value of mentioned thresholds; and finding the optimum threshold values is so hard in the real application. Secondly, the concept of independency criterion in WOCCE was just limited to random and initial points in the same type individual clustering algorithms, whereas based on the independency definition in WOC, it can be defined in the term of mathematical independent random variables for all kinds of clustering algorithms. Thirdly, the time complexity of A3 is really high because it is the average of the APMM for all existed clusters in a partition. Since APMM is technically designed for comparing the similarity between a partition and a cluster, there is a wide range of common parts that are sequentially repeated in the A3 metric. Lastly, the WOCCE is only developed for unsupervised learning, while this framework can be also used for semi-supervised learning \cite{Alizadeh15, Yousefnezhad15}. Indeed, this paper introduces a new framework for WOC-based CES to solve the mentioned problems in the WOCCE.
\section{The Proposed Method}
\subsection{Definition}
Based on outlines of the WOC theory \cite{Surowiecki04,Baker08,Alizadeh15}, the conditions for a crowd to be wise are: diversity, independency, decentralization, and aggregation. Baker et al. \cite{Baker08} and Alizadeh et al. \cite{Alizadeh15} redefined the WOC criteria for supervised learning and unsupervised learning, respectively. They used algorithms, data and results instead of people, information and opinions in the mentioned definitions, respectively. Same structure is utilized in this paper to redefine the criteria for proposing a framework in both unsupervised and semi-supervised methods. So, our definition for WOC criteria listed as follows:\\
\emph{\textbf{Independency}} The data, which is applied to clustering models, must have the lowest correlation between its features.\\
\emph{\textbf{Decentralization}} Algorithms are able to specialize the results based on the local knowledge. \\
\emph{\textbf{Diversity}} Each algorithm has private result, even if it is only an eccentric interpretation of the known facts.\\
\emph{\textbf{Aggregation}} Some method exists for combining private results into a collective decision (final result).

As a whole, it can be stated that the WoCE can produce final results in four stages. Firstly, the mapping function removes the correlation between the features of raw data set. This mapping function can satisfy independency criterion. Then, for satisfying the decentralization criterion, this paper applies local knowledge, i.e. the given number of clusters and supervision information. Further, it employs the various kinds of individual clustering algorithms. After these steps, diversity criterion evaluates the probability of accuracy in the generated clustering results. Finally, an effective aggregation method can increase the performance of the proposed method. In the rest of this section, the formulation of the proposed method will be discussed, and this paper will mention what WOC criterion is satisfied by using each part of the formulation. After that, we briefly summarized the whole algorithm procedure.
\subsection{Independency}
Based on definitions of the WOC theory, people must decide by using independent information. Hence, people can discover novel patterns, which are utilized to solve complex problems such as selecting the best person in the presidential election or finding an irregular engineering problem in the NASA's shuttle \cite{Surowiecki04,Alizadeh15}. In machine learning arena, this concept can be defined in the term of mathematical independent random variables. In fact, independent features are generated by removing the correlation between the features of raw data. There are various methods for removing the correlation before applying individual clustering techniques, such as Principle Component Analysis (PCA) or Linear Discriminant Analysis (LDA).  They can validate that removing the correlation dramatically improves the performance of clustering results \cite{Tan05,Yousefnezhad15}. Now, this paper defines independency criterion by utilizing the concept of correlation. In other words, this paper develops a new branch of mentioned methods in the CES for mapping data to different dimensions with less correlation between its features. In the rest of this section, we show that how our proposed method transforms features of raw data to stable dimensions with less correlation.

Given a set of data examples $\hat{X} = \{\hat{x}_{1},\hat{x}_{2},\ldots,\hat{x}_{n}\}$, and the corresponding pairwise must-link constraint set  $M= \{\left({x}_{i},{x}_{j}\right)$; ${x}_{i}$ and ${x}_{j}$  belong to the same cluster$\}$ and pairwise cannot-link constraint set $C=\{\left({x}_{i},{x}_{j}\right)$; ${x}_{i}$  and ${x}_{j}$  belong to different clusters$\}$. The simple average of $\hat{X}$ can be denoted as follows:
\begin{equation}
\bar{X} = \frac{1}{n}\sum_{i=1}^{n}\hat{x}_{i}
\end{equation}
where $n$ is the number of instances in the $\hat{X}$; and $\hat{x}_{i}$ denotes the $i-th$ instance of the data points. Now, this paper denotes $X$ as follows:
\begin{equation}
	X = \hat{X} - \bar{X} = \{(\hat{x}_{1} - \bar{x}_{1}), (\hat{x}_{2} - \bar{x}_{2}), \ldots, (\hat{x}_{n} - \bar{x}_{n})\}
\end{equation}
where $\hat{X}$ is the data points, and $\bar{X}$ denotes simple average of $\hat{X}$, which calculated by (1). It's clear that $X$ is zero-mean. In other words, the excepted value of $X$ is zero as follows:
\begin{equation}
\mathbb{E}\{X\}=0
\end{equation}
Further, this paper defines $Q: X\in \mathbb{R}^{m\times n} \to Y\in \mathbb{R}^{m\times n}$, where $m$, $n$ denote the number of features and data points, respectively. The main goal of this mapping is just minimizing the correlation between features. This problem can be reformulate as follows:
\begin{equation}
Y = {Q}^{\top}X
\end{equation}
If the correlation (covariance) of $X$ is considered $R = \mathbb{E}\{X{X}^{\top}\} = \frac{1}{n}\sum_{i = 1}^{n}{x}_{i}{x}^{\top}_{i} $, then the correlation of $Y$ will be defined as follows:
\begin{equation}
\begin{multlined}
\mathbb{E}\{Y{Y}^{\top}\} = \mathbb{E}\{({Q}^{\top}X){({Q}^{\top}X)}^{\top}\} =\\ 
\mathbb{E}\{{Q}^{\top}X{X}^{\top}Q\} = {Q}^{\top}\mathbb{E}\{X{X}^{\top}\}Q = {Q}^{\top}RQ
\end{multlined}
\end{equation}
Based on above definition, the expected value of $j-th$ feature of $X$ denotes as follows:
\begin{equation}
\mathbb{E}\{{Y}_{j}{Y}^{\top}_{j}\} = {q}^{\top}R{q}
\end{equation}
where $q$ denotes the $j-th$ index of the $Q$. In other words, our correlation problem is changed to a variance problem. Now, maximizing the $q$ based on the variance of $X$ will be omitted the correlation between features. Since the scale of data after mapping must be same, we assume following equation:   
\begin{equation}
\|q\| = 1. 
\end{equation}
So, our problem will be reformulated as follows:
\begin{equation}
\begin{multlined}
max[\Psi(q) = {q}^{\top}Rq] \Rightarrow\\
\frac{\partial\Psi(q)}{\partial q} = 0 \Rightarrow\\
\Psi(q + \delta q) = \Psi(q) \Rightarrow\\
{(q + \delta q)}^{\top}R(q + \delta q) = {q}^{\top}R{q}
\end{multlined}
\end{equation}
where the symbol $\delta q$ is an abbreviation for `a small change in q'. We consider $(\delta q)^{\top}\delta q \approx 0,$ so the above definition denotes as follows:
\begin{equation}
(\delta q)^{\top}Rq = 0
\end{equation}
Based on (7) and (8), we can assume as follows:
\begin{equation}
\|\delta q - q\| = \|q\| = 1 \Rightarrow (\delta q)^{\top}q = 0
\end{equation}
Now, this paper defines following equation by using (9) and (10):
\begin{equation}
\begin{multlined}
(\delta q)^{\top}Rq - \lambda(\delta q)^{\top}q = 0 \Rightarrow\\
(\delta q)^{\top}[Rq - \lambda q] = 0
\end{multlined}
\end{equation}
where $\lambda \in \mathbb{R}$ is a constant. Since $(\delta q)^{\top} \neq 0,$ the following equation must be satisfy for minimizing correlation between features:
\begin{equation}
Rq = q \lambda
\end{equation}
where R and $\lambda$ denote the eigenvectors and eigenvalues, respectively. For all features of $X$ the above equation will be denoted as follows:
\begin{equation}
RQ = Q \Lambda
\end{equation}
which is called eigenstructure equation. In above equation, $\Lambda$ is a diagonal matrix. Based on (7), we can define following equation:
\begin{equation}
\|q\|^{2} = 1 \Rightarrow {Q}^{\top}Q = \mathbb{I}
\end{equation}
where $\mathbb{I}$ is identity matrix. Following equation denotes based on (13) and (14):
\begin{equation}
\begin{multlined}
RQ = Q \Lambda \Rightarrow\\
RQ{Q}^{\top} = Q \Lambda{Q}^{\top} \Rightarrow\\
R\mathbb{I} = Q \Lambda{Q}^{\top} \Rightarrow\\
R = {Q}^{\top}\Lambda Q \Rightarrow\\
R = \sum_{j = 1}^{m} \lambda_{j} {q}_{i} {q}^{\top}_{j}
\end{multlined}
\end{equation}
where $m$ denotes number of features in data $X$. Now, consider that $R$ is a descending order based on $\Lambda$ values. For an optional feature selection in our unsupervised approach, we can define the following equation instead of (15):
\begin{equation}
R = \sum_{j = 1}^{d} \lambda_{j} {q}_{i} {q}^{\top}_{j}
\end{equation}
where $d < m$ is the number of features, which must be selected for generating results. Algorithm \ref{alg:Mapping} shows the mapping function, which can generate independent features by minimizing the correlation of data set. For reducing the time complexity, this paper uses an EM algorithm \cite{tipping99} for estimating the eigenvalues/vectors ($\Lambda$ and $Q$) in Algorithm \ref{alg:Mapping}. Please see Section A.5 in \cite{tipping99} for more information. 
\begin{algorithm}[H]
	\caption{The Mapping Function}
	\label{alg:Mapping}
	\begin{algorithmic}
		\STATE {\bfseries Input:} Data set $\hat{X} = \{\hat{x}_{1},\hat{x}_{2},\ldots,\hat{x}_{n}\}$,\\
		\quad $d$ as number of features: \\ 
		\quad \emph{$d = 0$ is considered for deactivating the feature selection}
		\STATE {\bfseries Output:} Mapped data set $Y$
		\STATE {\bfseries Method:}\\
		\quad1. Calculate simple average $\bar{X}$ by using (1).\\
		\quad2. Calculate $X$ by using (2).\\
		\quad3. Generate $R = \mathbb{E}\{X{X}^{\top}\} = \frac{1}{n}\sum_{i = 1}^{n}{x}_{i}{x}^{\top}_{i}$.\\
		\quad4. Calculate eigenvalues/vectors ($\Lambda$ and $Q$) of $R$ by \cite{tipping99}.\\
		\quad5. Sort $Q$ based on descending values of $\lambda$. \\
		\quad6. \textbf{if} d is not zero ($d \neq 0$) \textbf{then}\\
		\quad\qquad Select $[1, d]$ features of $Q$, and sorting as ${Q}_{d}$,\\ 
		\quad\quad \textbf{else} ${Q}_{d} = Q$.\\ 
		\quad\quad \textbf{end if}\\
		\quad7. Return $Y = {Q}_{d}^{\top}X$.\\ 
	\end{algorithmic}
\end{algorithm}
\subsection{Decentralization}
In WOC theory, the decentralization criterion increases the crowd intelligence, the margin of error and the quality of the final result \cite{Surowiecki04,Alizadeh15}. In the clustering problems, the same concept is the main reason for using the CES approach to improve the quality of the final result. So, there is a wide range of quality metrics in the previous CES methods \cite{Fern08,Alizadeh12,Alizadeh14}. Based on the WOC theory, this paper uses local knowledge for increasing the quality of individual clustering results. There are two different kinds of local knowledge in the CES; i.e. the number of clusters in unsupervised learning and supervision information in semi-supervised learning. Moreover, employing different kinds of clustering algorithms significantly can affect to generate more specialize clustering results because they include different kinds of objective functions \cite{Alizadeh15}. Briefly, this paper applies the different kinds of clustering algorithms on the mapped data for generating the individual clustering results in both unsupervised and semi-supervised versions of the proposed method. Further, these algorithms use different numbers of clusters in the range of $[2, k+2]$, where k denotes the number of clusters in the final results. Since, this procedure generates all available kinds of patterns as the reference set, it can increase the robustness of the final results. In addition, this paper develops a new feature selection method based on supervision information for improving the performance of the final result. In the rest of this section, we show that how this paper uses supervision information for generating common/local knowledge in the semi-supervised approach.

As mentioned before, our proposed method is based on pairwise constraint, i.e. must-links and cannot-links. This paper denotes the must-link constraint with $M$, and the cannot-link constraint with $C$. For generating each individual clustering result, this paper defines Constraint Projection, which is a set of projective vectors $W = \left[{w}_{1},{w}_{2},\ldots,{w}_{d}\right]$, such that the $M$ and $C$ are most faithfully preserved in the transformed low-dimensional representations ${z}_{i}  = {W}^{\top}{y}_{i}$. That is, examples involved by $M$ should be close while examples involved by $C$ should be far in the low-dimensional space. Define the objective function as maximizing $J(W)$ with respect to ${W}^{\top}W=\mathbb{I}$, where: 
\begin{equation}
\begin{multlined}
\label{eq:J(W)}
J\left(W\right) = \frac{1}{2{n}_{C}}\sum_{\left({y}_{i},{y}_{j}\right)\in C}\|{z}_{i}-{z}_{j}{\|}^{2} \\ 
\quad - \frac{\gamma}{2{n}_{M}}\sum_{\left({y}_{i},{y}_{j}\right)\in M}\|{z}_{i}-{z}_{j}{\|}^{2}\\
\quad =\frac{1}{2{n}_{C}}\sum_{\left({y}_{i},{y}_{j}\right)\in C}\|{W}^{\top}{y}_{i}-{W}^{\top}{y}_{j}{\|}^{2} \\ 
\quad - \frac{\gamma}{2{n}_{M}}\sum_{\left({y}_{i},{y}_{j}\right)\in M}\|{W}^{\top}{y}_{i}-{W}^{\top}{y}_{j}{\|}^{2}
\end{multlined}
\end{equation}
where ${n}_{C}$ and ${n}_{M}$ denote the cardinalities of $C$ and $M$, respectively, and $\gamma$ is a scaling coefficient. The intuition behind (17) is to let the average distance in the low-dimensional space between examples involved by the cannot-link $C$ as large as possible, while distances between examples involved by the must-link $M$ as small as possible. Since the distance between examples in the same cluster is typically smaller than that in different clusters, a scaling parameter $\gamma$ is added to balance the contributions of the two terms in (17) and its value can be estimated as follows:
\begin{equation}
\label{eq:gamma}
\gamma = \frac{\frac{1}{{n}_{C}}\sum_{\left({y}_{i},{y}_{j}\right)\in C}\|{y}_{i}-{y}_{j}{\|}^{2}}{\frac{1}{{n}_{M}}\sum_{\left({y}_{i},{y}_{j}\right)\in M}\|{y}_{i}-{y}_{j}{\|}^{2}}
\end{equation}
We can also reformulate the objective function in (17) in a more convenient way as follows:
\begin{equation}
\label{eq:JW2}
J\left(W\right) = trace\left({W}^{\top}\left({S}_{C}-\gamma{S}_{M}\right)W\right)
\end{equation}
where ${S}_{C}$ and ${S}_{M}$ are respectively defined as:
\begin{equation}
\label{eq:Sc}
{S}_{C} = \frac{1}{2{n}_{C}}\sum_{\left({y}_{i},{y}_{j}\right)\in C}\left({y}_{i}-{y}_{j}\right)\left({y}_{i}-{y}_{j}\right)^{\top}
\end{equation}
\begin{equation}
\label{eq:Sm}
{S}_{M} = \frac{1}{2{n}_{M}}\sum_{\left({y}_{i},{y}_{j}\right)\in M}\left({y}_{i}-{y}_{j}\right)\left({y}_{i}-{y}_{j}\right)^{\top}
\end{equation}
This paper calls ${S}_{C}$ and ${S}_{M}$ defined in (20) and (21) respectively as cannot-link scatter matrix and must-link scatter matrix, which resemble the concepts of between-cluster scatter matrix and within-cluster scatter matrix respectively in linear discriminant analysis (LDA) \cite{Tan05}. The difference lies in that the latter uses cluster labels to generate scatter matrices, while the former uses pairwise constraints to generate scatter matrices. Obviously, the problem expressed by (19) is a typical eigen-problem, and can be efficiently solved by computing the eigenvectors of ${S}_{C}-\gamma{S}_{M}$  corresponding to the positive eigenvalues. In other words, just consider that $\bar{W}$ and $Z = \{{\zeta}_{1}, {\zeta}_{2}, \ldots, {\zeta}_{p}, \ldots, {\zeta}_{d}\}$ are eigenvectors and eigenvalues of ${S}_{C}-\gamma{S}_{M}$, respectively. The $\bar{W}$ and $\zeta$ is descending ordered based on $\zeta$ values (${\zeta}_{1} \geq {\zeta}_{2} \geq \ldots \geq  {\zeta}_{p} > 0 \geq \ldots \geq {\zeta}_{d}$). Also, $W = \{ \bar{W}$ $|$ $ \forall   \bar{W}_{p}$ where $p$ shows the position of positive eigenvalues (${\zeta}_{p} > 0)\}$. Further, the transformed data set is calculated as follows:
\begin{equation}
\label{eq:Transformed data}
Z = {W}^{\top}Y
\end{equation}
Algorithm 2 illustrates the transformation algorithm for both unsupervised and semi-supervised approaches. The transformed data is applied to different kinds of individual clustering algorithms for generating the reference set.
\begin{algorithm}
	\caption{The Transformation Algorithm}
	\label{alg:transforming}
	\begin{algorithmic}
		\STATE {\bfseries Input:} data set $\hat{X}$, \\
		\quad must-links $M$, cannot-links $C$, \\
		\quad\quad \emph{(as supervision information)}\\
		\quad Number of features $d$:\\ 
		\quad\quad \emph{(as optional feature selection)}
		\STATE {\bfseries Output:} Mapped data set $Z$
		\STATE {\bfseries Method:}\\
		\quad1. Generating Y by using Algorithm 1 and $\hat{X}$ and $d$.\\
		\quad2. \textbf{if} $M$ and $C$ are empty\\
		\quad \textbf{then} return $Z = Y$\\
		\quad \textbf{end if}\\
		\quad3. Generating ${S}_{M}$, ${S}_{C}$, $\gamma$ by using $Y$, (18), (20), (21)\\
		\quad4. Calculating the eigenvalues $\bar{W}$ and eigenvectors $\zeta$\\ \quad\quad of ${S}_{C}-\gamma{S}_{M}$.\\
		\quad5. Calculating the $W$ by using $\bar{W}_{p}$ based on $\zeta > 0$.\\ 
		\quad6. Return $Z = {W}^{\top}Y$\\
	\end{algorithmic}
\end{algorithm}

\subsection{Diversity}
Indeed, diversity is a common concept in both the WOC theory and the CES methods. For instance, NMI \cite{Fred05} and APMM \cite{Alizadeh14} are two famous methods for calculating diversity in the cluster ensemble (selection). The diversity increases the stability of the final results. As mentioned before, NMI has the symmetric problem. This problem causes that evaluation of the diversity between two clusters always results equal, when those clusters are complements of each other. This fault is occurred when the number of positive clusters in the considered partition of reference set is greater than 1 \cite{Alizadeh12,Alizadeh14,Alizadeh15}. Although some of the researches proposed alternative methods such as APMM \cite{Alizadeh14} and MAX \cite{Alizadeh12} for solving this problem, their proposed methods were utilized for evaluating diversity between a cluster and a partition. As a result, using mentioned methods for evaluating the diversity of two partitions increases the time complexity. In the rest of this section, we firstly explain that how NMI and APMM work. Then, we develop a new metric, which directly can evaluate diversity between two partitions.

Indeed, NMI employed three different Shannon's entropy for evaluating the similarity between two partitions. Since, NMI is normalized, the $1 - NMI$ was always considered as the diversity between mentioned partitions. NMI used the entropy of common instances between two partitions as numerator, and also employed the sum of entropy of each partition as denominator \cite{Alizadeh12,Alizadeh14,Fred05}. As mentioned before, NMI has symmetric problem. As another alternative, APMM tried to solve the mentioned problem for evaluating the similarity between a cluster ($C_i^a$ from $P^a$) and all clusters of another partition ($P^b$) \cite{Alizadeh14}. Since, some common parts of APMM must be repeated for calculating diversity of two partitions, using the APMM for evaluating the diversity of two partitions increases the time complexity. Further, simple average was utilized for calculating the diversity between all clusters of a partition ($P^a$) versus all clusters of another partition ($P^b$) \cite{Alizadeh14,Alizadeh15}. This averaging procedure causes to decrease the robustness of achieved evaluation because it finds the mean of similarity between all clusters of two partitions instead of calculating maximum similarity (minimum diversity) among of them. This paper proposes a new greedy method based on the main idea of the APMM. It can calculate diversity between two partitions without repeating common parts; and also it avoids using the averaging procedure.

As mentioned in the previous section, individual clustering results are generated by using the transformed data on the different kinds of clustering algorithms. This paper denotes the generated results as a reference set as follows:
\begin{equation}
\label{eq:Reference Set}
E = \{{P}_{1}, {P}_{2}, \ldots, {P}_{i}, \ldots, {P}_{T}\}
\end{equation}
where $T$ denotes the number of individual clustering results and ${p}_{i}$ is the $i-th$ partition of the generated results. Now, this paper finds the maximum similarity for each partition by considering the number of all instances in that partition versus the number of instances in each cluster of that partition as follows: 
\begin{equation}
\label{eq:Entropy(N/Nc)}
\eta(P)={\max}_{{c}_{i} \in {P}}({n}_{i}\log(\frac{n}{{n}_{i}})) 
\end{equation}
where $P$ is a partition from the reference set; ${c}_{i}$ denotes the $i-th$ cluster of partition $P$, and $n$ and ${n}_{i}$ denote the cardinalities of $P$ and ${c}_{i}$, respectively. Furthermore, this paper finds the maximum similarity for each partition by considering  the number of instances in each cluster of that partition versus the number of all instances in that partition as follows: 
\begin{equation}
\label{eq:Entropy(Nc/N)}
\xi(P)={\max}_{{c}_{i} \in {P}}({n}_{i}\log(\frac{{n}_{i}}{n})) 
\end{equation}
where the notations of $P$, ${c}_{i}$, $n$, ${n}_{i}$ define same as the previous equation. Now, this paper determines the following equation as maximum similarity between a partition versus other partitions in the reference set:
\begin{equation}
\label{eq:Entropy(P/E)}
\Theta(P,E)={\max}_{{P}_{i} \in {E}}(\max_{{c}_{j} \in {P}_{i}}{n}_{i}^{j}\log(\frac{{n}_{i}^{j}}{n}))
\end{equation}
where $E$ and $P$ are the reference set and a partition from the reference set, respectively. Also, ${P}_{i}$ and ${c}_{j}$ denote the $i-th$ partition from the reference set $E$ and $j-th$ cluster from the partition ${P}_{i}$, respectively. Further, ${n}_{i}^{j}$ and $n$ are the cardinalities of ${c}_{j}$ and $P$, respectively. Now, this paper proposes the Uniformity as the diversity of partition $P$ versus all partitions of the reference set $E$ as follows:
\begin{equation}
\begin{multlined}
\label{eq:Uniformity}
Uniformity(P, E)= 1 - \frac{-2\eta(P)}{\xi(P) + \Theta(P,E)}
\end{multlined}
\end{equation}
where $E$ is the reference set (ensemble committee), and $P$ denotes a partition from the reference set. Uniformity is normalized between $0\le Uniformity\le 1$. As a greedy metric, Uniformity employs a strict strategy for evaluating the diversity between partition $P$ and the other partitions of ensemble committee. In other words, Uniformity represents a value near of zero for a partition with low diversity, and illustrates a value near of one for a partition with high diversity. In addition, it avoids to repeat common parts, i.e. equations (25) and (26), for evaluating the diversity in each comparison.
\subsection{Aggregation}
Thresholding is used for selecting the evaluated individual results in the CES. Then \textit{co-association} matrix is generated by using consensus function on the selected results. Lastly, the final result  is generated by applying linkage methods on the co-association matrix. These methods generate the Dendrogram and cut it based on the number of clusters in the result \cite{Fred05,Alizadeh15}. In recent years, many papers have used EAC as a high-performance consensus function for combining individual results \cite{Fred05,Alizadeh14,Alizadeh15,Azimi09,Fern08}. EAC uses the number of clusters shared by objects over the number of partitions in which each selected pair of objects is simultaneously presented for generating each cell of the co-association matrix.
\begin{figure}[!t]
	\centering
	\includegraphics[width=2.5in]{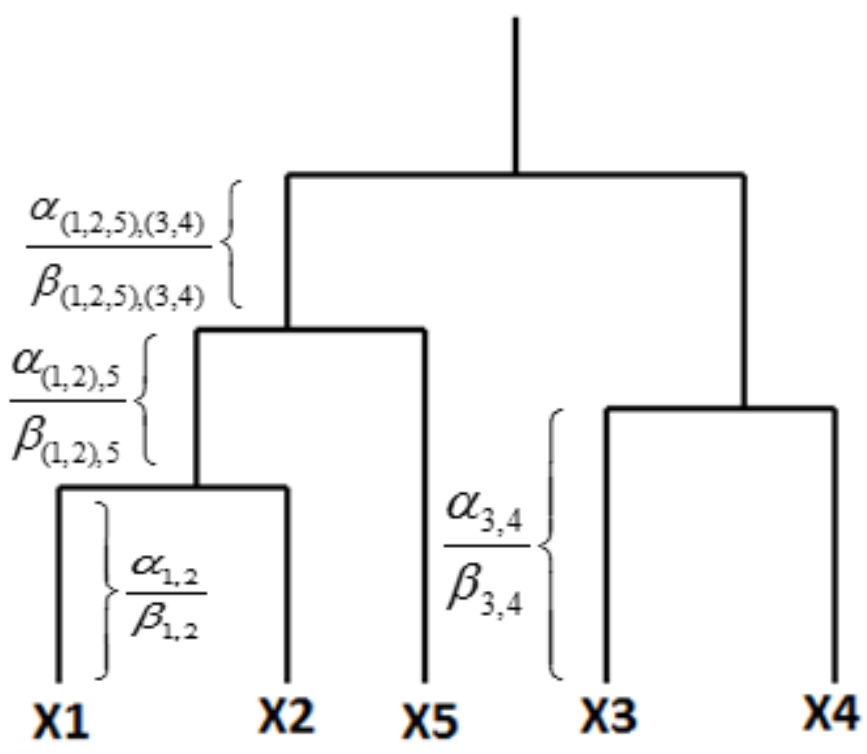}
	\caption{
		In the traditional EAC, the ${\alpha}_{(i,j)}$ represents the number of clusters shared by objects with indices (i, j); and ${\beta}_{(i,j)}$ is the number of partitions in which this pair of instances (i and j) is simultaneously presented. This method assumes the weights of all individual clustering results (${\alpha}_{(i,j)}$) are the same. This paper proposes Weighted EAC for optimizing this method by using a weight for each individual clustering results instead of just counting their shared clusters. While the weight can have different definitions in the other applications, this paper uses average of Uniformity of two parititon as the weight in the WEAC ($\bar{{\alpha}}_{(i,j)}= \sum_{\alpha\left(i,j\right)}{\rho}_{i,j}$).}
	\label{Fig1}
	\vskip -0.2in
\end{figure}
Figure 1 illustrates the effect of the EAC equation ($c\left(i,j\right)=\frac{\alpha\left(i,j\right)}{\beta\left(i,j\right)}$) on the shape of Dendrogram. Where ${\alpha}_{(i,j)}$ represents the number of clusters shared by objects with indices (i, j); and ${\beta}_{(i,j)}$ is the number of partitions in which this pair of instances (i and j) is simultaneously presented. As a matter of fact; EAC considers that the weights of all algorithms’ results are the same. Instead of counting these indices, this paper uses following equation, which is called Weighted EAC (WEAC), for generating the co-association matrix.
\begin{equation}
\label{WEAC}
c\left(i,j\right)=\frac{\sum_{\alpha\left(i,j\right)}{\rho}_{i,j}}{\beta\left(i,j\right)}
\end{equation} 
where $\alpha\left(i,j\right)$ and $\beta\left(i,j\right)$ are same as the EAC equation; Also, ${\rho}_{i,j}$ is the weight of combining the instances. Although, this weight can have different definitions in the other applications, this paper uses average of Uniformity of two algorithms as follows for combining individual results:  
\begin{equation}
\label{AvrageOfUniformity}
{\rho}_{ij}=\frac{1}{2}(Uniformity({P}_{i},E) + Uniformity({P}_{j},E))
\end{equation}   
where $Uniformity({P}_{i},E)$ and $Uniformity({P}_{j},E)$ illustrate the uniformities of the algorithms, which generated the results for indices $i$ and $j$. In other words, as a new mechanism, this paper generates the effective results when both algorithms have high Uniformity values; and also the effects of individual results are near of zero when the both algorithms have small values in the Uniformity metric. As a result, this paper just omits the effect of low quality individual results by using mentioned mechanism instead of selecting them by thresholding procedures. Further, the final co-association matrix, which is a symmetric matrix, will be generated by (28) as follows:
\begin{equation}
\label{co-association matrix}
\Pi = WEAC(\tau) = \left( 
\begin{array}{cccc}
c(1,1) & c(1,2) & \dots & c(1,n) \\
c(2,1) & c(2,2) & \dots & c(2,n) \\
\vdots & \vdots & \vdots & \vdots \\
c(i,1) & c(i,2) & c(i,j) & c(i,n) \\
\vdots & \vdots & \vdots & \vdots \\
c(n,1) & c(n,2) & \dots & c(n,n) \\
\end{array} \right)
\end{equation}   
where $n$ is the number of data points; and $c(i,j)$ denotes the final aggregation for $i-th$ and $j-th$ instances.
\begin{algorithm}[H]
	\caption{The WoCE algorithm}
	\label{alg:SCEWOC}
	\begin{algorithmic}
		\STATE {\bfseries Input:} Data set $\hat{X}=\{\hat{x}_{i}\}_{i=1}^{n}$,\\
		\quad Must-links $M$,\\
		\quad Cannot-links $C$,\\
		\quad Number of clusters $k$,\\
		\quad Number of selected features $d$ (default d=0),\\
		\STATE {\bfseries Output:} ${P}_{f}$ as partition of data set into $k$ clusters 
		\STATE {\bfseries Method:}\\
		\quad1. Initial an empty set as Reference-Set.\\
		\quad2. Generate $Z$ by applying Algorithm 2 to ($\hat{X}$, $M$, $C$, $d$).\\
		\quad \textbf{foreach} individual clustering algorithms \textbf{do}\\
		\quad3. \quad $iResult$ = Clustering-Algorithm($Z$, $k$).\\
		\quad4. \quad diversity = Uniformity ($iResult$, $Reference-Set$).\\
		\quad5. \quad Add $\left[iResult, diversity\right]$ to $Reference-Set$\\
		\quad \textbf{end foreach}  \\
		\quad6. $Co-Association$ = WEAC($Reference-Set$)\\
		\quad7. Dendrogram = Average-Linkage ($Co-Association$)\\
		\quad8. Final-Result = Cluster ($Dendrogram$, $k$)
	\end{algorithmic}
\end{algorithm}
\subsection{Summarization and Discussion}
Algorithm 3 shows the pseudo code of the proposed method. In this algorithm, the distances are measured by an Euclidean metric. The Clustering-Algorithm function builds the partitions of individual clustering results, which will be discussed in the next section; uniformity function evaluates individual clustering results $\left(iResult\right)$ by using (27). Then, evaluated results will be added to reference set. The WEAC function generates the co-association matrix, according to (28). The Average-Linkage function creates the final ensemble according to the Average Linkage method \cite{Alizadeh15}. 

There are three points, which must be discussed before this paper starts to explain the empirical studies. Firstly, why this paper chooses the WOC as a framework in the cluster ensemble? As mentioned before, the main reasons for using cluster ensemble are increasing performance, stability, robustness of the final results on the clustering problems. As already stated, the WOC theory is superior to that of a few experts. In other words, it is proven \cite{Surowiecki04,Baker08,Alizadeh15} that results made by aggregating the information of groups have better performance, stability, and robustness than those made by any single group member if the WOC criteria are satisfied. Therefore, the cluster ensemble and the WOC are the same solutions with the same goals in two different sciences, i.e. machine learning and social science, respectively. Next, what are the common concepts between our proposed criteria in the WOC and previous methods in clustering problems? In fact, diversity is existed in clustering with the same title, e.g. NMI and APMM are two famous methods for calculating diversity in the cluster ensemble (selection). The diversity increases the stability and robustness of the final results. Further, independency referred to the correlation concept in the learning methods. This correlation can be defined between features of raw data. There are some techniques, i.e. Principle Component Analysis (PCA), Linear Discriminant Analysis (LDA), etc. for mapping data to new dimensions without any correlation between its features. This paper uses a new branch of these techniques for satisfying the independency criterion, which can increase the performance of the final results. In addition, decentralization guarantees that the quality of the final result is optimized. In other words, it uses different individual clustering algorithms, which use different objective functions, for generating all possible patterns as the reference set in the cluster ensemble problem. Moreover, an effective aggregation method can combine the final result without thresholding procedure. The last question is why all of the four conditions of the WOC must be satisfied in the ensemble learning? Based on previous question, the proposed method can be defined as a CES method that applied a feature mapping in advance. In practice, all of clustering analysis has these steps \cite{Yousefnezhad15,Alizadeh15,Gao13}. Therefore, the WOC framework does not add any new stage in the pipeline on clustering analysis. It just defined what is the robust and compulsory structure for an ensemble framework in real-world application.
\section{Experiments}
The empirical studies will be presented in this section. The unsupervised methods are used to find meaningful patterns in unlabeled data sets such as web documents; and semi-supervised employs supervision information for generating more robust and stable final results in real world application. Since, the real data set does not have class labels, there is no direct evaluation method for estimating the performance in unsupervised or semi-supervised methods. Like many previous researches \cite{Fern08,Alizadeh12,Alizadeh14,Alizadeh15,Gao13,Anand14}, this paper compares the performance of its proposed method with other individual clustering methods and cluster ensemble (selection) methods by using standard data sets and their real classes. Moreover, the supervision information will be randomly generated based on real class labels. In this paper, all of algorithms are implemented in the MATLAB R2015a (8.5) by authors on a PC with certain specifications\footnote{Apple Mac Book Pro, CPU = Intel Core i7 (4*2.4 GHz), RAM = 8GB, OS = OS X 10.11} in order to generate experimental results. All results are reported by averaging the results of 10 independent runs of the algorithms. Table I demonstrates the individual clustering algorithms, which are used for generating the individual clustering results in our proposed method. Further, the number of individual clustering results for the ensemble methods is set as 20 in the reference set. 
\begin{table}
	\caption{The individual clustering algorithms, which are used for generating individual clustering results}
	\label{Tbl1: Individiual clustering results}
	%\vskip 0.15in
	\begin{center}
		\begin{small}
			\rowcolors{1}{light-gray}{white}
			\begin{tabular}{cl}
				\hline
				No. & Method \\
				\hline
1   & K-means\\
2   & Fuzzy C-means \\
3   & Median K-flats\\
4   & Gaussian mixture\\
5   & Subtract clustering\\
6   & Single-linkage euclidean\\
7   & Single-linkage hamming\\
8   & Single-linkage cosine\\
9   & Average-linkage euclidean\\
10 & Average-linkage hamming\\
11 & Average-linkage cosine\\
12 & Complete-linkage euclidean\\
13 & Complete-linkage hamming\\
14 & Complete-linkage cosine\\
15 & Ward-linkage euclidean\\
16 & Ward-linkage hamming\\
17 & Ward-linkage cosine\\
18 & Spectral using a sparse similarity matrix\\
19 & Spectral using Nystrom method with orthogonalization\\
20 & Spectral using Nystrom method without orthogonalization\\
				\hline
			\end{tabular}
		\end{small}
	\end{center}
	\vskip -0.1in
\end{table}
\subsection{Data Sets}
This paper uses three different groups of data sets for generating experimental results; i.e. image data sets, document data sets and other UCI data sets. Table II illustrates the properties of these data sets. This paper uses the USPS digits data set, which is a collection of $16\times16$ gray-scale images of natural handwritten digits and is available from \cite{USPS}. Furthermore, this paper utilizes ImageNet \cite{Lazebnik06}, MNIST, and CIFAR-10 \cite{Azadi16} as three image-based data sets, which are mostly employed in Deep Learning studies \cite{Azadi16}. As another alternative in the image-based data set, this paper uses Alzheimer's Diseases Neuroimaging Initiative (ADNI) data set for $202$ subjects. This data set contains Magnetic Resonance Imaging (MRI) and Positron Emission Tomography (PET) images from human Brian in two categories (which are shown by C1 and C2 in the Table II and III) for recognizing the Alzheimer diseases. In the first category, this data set partitions subjects to three groups of Health Control (HC), Mild Cognitive Impairment (MCI), and Alzheimer's Diseases (AD). In the second category, there are four groups because the MCI will be partitioned into high and low risk groups (HMCI/LMCI). This paper uses all possible forms of this data set by using only MRI features, only PET features and all of MRI and PET features (FUL) in each of two categories. More information about ADNI-202 is available in \cite{Zu14}. As a document based data set, the 20 Newsgroups is a collection of approximately 20,000 newsgroup documents, which is partitioned (nearly) evenly across 20 different newsgroups. Some of the newsgroups are very closely related to each other, while others are highly unrelated. It has become a popular data set for experiments in text applications of machine learning techniques, such as text classification and text clustering. As two other document-based data sets, the Reuters-21578 \cite{cai11} and Letters \cite{Huang15} are employed in this paper. The rest of standard data sets are from UCI \cite{UCI}. This paper has chosen data sets which are as diverse as possible in their numbers of true clusters, features, and samples because this variety better validates the obtained results. The features of the data sets are normalized to a mean of 0 and variance of 1, i.e. $\mathcal{N}(0, 1)$.
\begin{table}
	\caption{The standard data sets}
	\label{Tbl2: Data Set}
	\vskip -0.5in
	\begin{center}
		\begin{small}
			\rowcolors{1}{light-gray}{white}
			\begin{tabular}{lccc}
				\hline
				Data Set & Instances & Features & Class \\
				\hline
				20 Newsgroups & 26214 & 18864 & 20 \\
				ADNI-MRI-C1  & 202  & 93 & 3 \\
				ADNI-MRI-C2  & 202  & 93 & 4 \\
				ADNI-PET-C1  & 202  & 93 & 3 \\
				ADNI-PET-C2  & 202  & 93 & 4 \\
				ADNI-FUL-C1  & 202 & 186 & 3 \\
				ADNI-FUL-C2  & 202 & 186 & 4 \\		
				Arcene          &  900 & 10000 & 2  \\
				Bala. Scale    &  625 & 4 & 3 \\
				Brea. Cancer &  286 & 9 &  2 \\
				Bupa             & 345 & 6 & 2 \\ 
				CIFAR-10       & 5000 & 1024 & 10 \\
				CNAE-9         & 1080 & 857 & 9 \\
				Galaxy           & 323 & 4 & 7 \\
				Glass             &  214 & 10 & 6 \\
				Half Ring        &  400 & 2 & 2 \\
				ImageNet    & 5000 & 400 & 5 \\
				Ionosphere    & 351 & 34 & 2 \\
				Iris                 & 150 & 4 & 3 \\
				Letters		  & 20000 & 16 & 26 \\
				MNIST           & 70000 & 784 & 10 \\
				Optdigit          & 5620 & 62 & 10 \\
				Pendigits       & 10992 & 16 & 10 \\
				Reuters-21578 & 8293 & 18933 & 65\\
				SA Hart         & 462 & 9 & 2 \\
				Sonar            & 208 & 60 & 2 \\
				Statlog           & 6435 & 36 & 7 \\
				USPS             & 9298 & 256 &  10 \\
				Wine              & 178 & 13 & 2 \\
				Yeast              & 1484 & 8 & 10 \\
				\hline
			\end{tabular}
			%			\end{sc}
		\end{small}
	\end{center}
	\vskip -0.3in
\end{table}
\subsection{Performance analysis for unsupervised methods}
In this section the performance (accuracy metric \cite{Tan05}) of unsupervised version of proposed method (UWoCE) will be analyzed. As mentioned before, algorithms listed in Table \ref{Tbl1: Individiual clustering results} were employed for generating the individual clustering results in our proposed method. Further, the sets of supervision information (must-links and cannot links) are considered null in this section. Also, the final clustering performance was evaluated by re-labeling between obtained clusters and the ground truth labels and then counting the percentage of correctly classified samples \cite{Alizadeh15,Yousefnezhad15}. The results of the proposed method are compared with full ensemble (EAC) \cite{Fred05} as baseline, WPCK \cite{Vega10}, GKPC \cite{Vega11}, HCSS \cite{Yu14}, GP-MGLA \cite{Huang15}, and WOCCE \cite{Alizadeh15} which are state-of-the-art cluster ensemble (selection) methods. The performance of full ensemble method (EAC) is reported for demonstrating the effect of selecting the best results in comparison combing all generated result with each others. In addition, the performance of the WPCK, GKPC, HCSS, and GP-MGLA are reported as four weighted clustering ensemble methods. For representing the effect of Uniformity on the performance of the final results, it compares with three state-of-the-art metrics in diversity evaluation (A3 \cite{Alizadeh15}, SACT \cite{Huang15}, and CA \cite{Vega11}). This paper does not use optional feature selection in this section ($d = 0$). 

The experimental results are given in Table \ref{Tbl3: Unsupervised Expermintal Results}. In this table, the best result which is achieved for each data set is highlighted in bold. As depicted in this table, the results of the EAC illustrate the effect of evaluation and selection in cluster ensemble selection methods. Since some of the four conditions of the WOC theory do not exist in EAC, this method is a good example of unwise crowd. According to this table, the proposed algorithm (WoCE) has generated better results in comparison with other individual and ensemble algorithms. Even though the proposed method was outperformed by a number of algorithms in four data sets (ADNI-MRI-C2, SA Heart, Sonar, and Yeast), the majority of the results demonstrate the superior accuracy of the proposed method in comparison with other algorithms. In addition, the difference between the performance of proposed method and the best result in those three data sets is lower that 2\%. In addition, the WOCCE and the proposed method generate more stable results in comparison with other methods based on the standard variances. As mentioned before, this is the effect of WOC framework.
\begin{table*}
\caption{The performance of unsupervised methods}
\label{Tbl3: Unsupervised Expermintal Results}
\begin{center}\begin{small}\begin{sc}
\rowcolors{1}{light-gray}{white}
\begin{tabular}{lccccccc}
\hline
Data Sets & EAC & WPCK & GKPC & HCSS & GP-MGLA & WOCCE  & UWoCE \\
\hline
20 Newsgroups & 26.19$\pm$0.72 & 27.01$\pm$0.93  & 28.45$\pm$1.02 & 30.62$\pm$0.84  & 35.47$\pm$0.91& 32.62$\pm$0.52   & \textbf{38.23$\pm$0.12}  \\
ADNI-MRI-C1  &   42.19$\pm$0.37 & 41.24$\pm$0.97  & 43.51$\pm$1.02 &  46.61$\pm$0.36  & 49.36$\pm$0.7& 48.82$\pm$0.37 &  \textbf{51.15$\pm$0.73}\\
ADNI-MRI-C2  & 39.52$\pm$0.31 & 39.95$\pm$0.61  & 40.09$\pm$0.51 & 41.32$\pm$0.81  & 40.72$\pm$1.25& \textbf{42.22$\pm$0.44} & 41.23$\pm$0.95\\
ADNI-PET-C1  & 40.38$\pm$0.52 & 40.51$\pm$0.26  & 43.79$\pm$1.04& 45.3$\pm$0.49  & 48.22$\pm$0.71& 49.19$\pm$0.26   &  \textbf{51.17$\pm$0.98}\\
ADNI-PET-C2 & 38.85$\pm$0.59 & 37.51$\pm$0.69  & 36.58$\pm$0.72& 41.92$\pm$1.18  & 40.68$\pm$0.73& 39.43$\pm$0.79  & \textbf{42.48$\pm$0.67}\\
ADNI-FUL-C1  & 44.42$\pm$0.91 & 43.84$\pm$0.93  & 46.56$\pm$0.49 & 49.62$\pm$0.81  & 49.27$\pm$0.61& 48.82$\pm$0.41& \textbf{50.89$\pm$0.83}\\
ADNI-FUL-C2 & 47.21$\pm$0.63 & 49.71$\pm$0.99  & 51.26$\pm$0.64 & 52.26$\pm$0.66  & 51.92$\pm$0.7& 49.39$\pm$0.63 &  \textbf{53.31$\pm$0.97}\\									
Arcene & 61.79$\pm$0.813& 63.92$\pm$0.81  & 63.26$\pm$1.04 & 65.54$\pm$0.73  & 66.32$\pm$0.91& 65.16$\pm$0.32 &  \textbf{68.13$\pm$0.82} \\
Bala. Scale  & 54.09$\pm$1.75 & 55.42$\pm$0.94  & 56.04$\pm$0.72 & 57.41$\pm$0.56  & 56.23$\pm$0.94& 57.88$\pm$0.61 &  \textbf{60.64$\pm$0.58} \\
Brea. Cancer  & 90.17$\pm$1.24& 81.93$\pm$1.92  & 82.43$\pm$1.24 & 65.51$\pm$1.91  & 72.27$\pm$1.06& 96.92$\pm$0.77 &   \textbf{97$\pm$0.14} \\
Bupa & 51.73$\pm$0.99 & 57.91$\pm$0.82  & 59.09$\pm$0.98 & 58.33$\pm$1.32  & 58.91$\pm$0.51& 57.02$\pm$0.46 & \textbf{60.83$\pm$0.12} \\
CIFAR-10  & 51.92$\pm$1.24& 54.1$\pm$0.88  & 55.52$\pm$0.79  & 56.12$\pm$0.91& 57.82$\pm$0.85  & 59.37$\pm$0.52&  \textbf{62.04$\pm$0.32} \\
CNAE-9  & 72.41$\pm$1.09 & 75.41$\pm$0.69  & 75.53$\pm$0.55& 80.63$\pm$1.41  & 81.29$\pm$0.81& 79.2$\pm$0.58&  \textbf{84.12$\pm$0.44} \\
Galaxy    & 33.12$\pm$0.52& 30.99$\pm$1  & 32.71$\pm$0.84 & 35.71$\pm$0.61  & 34.72$\pm$0.96& 34.88$\pm$0.81 &  \textbf{37.18$\pm$0.67} \\
Glass     & 50.93$\pm$0.18& 45.01$\pm$2.03  & 46.57$\pm$2.97 & 52.31$\pm$0.68  & 50.62$\pm$0.38& 51.82$\pm$0.92 &   \textbf{57$\pm$0.78}\\
Half Ring & 77.53$\pm$0.21& 82.54$\pm$0.93  & 85.41$\pm$0.94 & 90.53$\pm$0.67  & 89.99$\pm$1.02& 87.2$\pm$0.14 &  \textbf{98.11$\pm$0.31}\\
ImageNet   & 23.53$\pm$0.81& 32.86$\pm$0.42  & 35.32$\pm$0.59& 35.04$\pm$0.93& 33.51$\pm$0.83  & 38.14$\pm$0.62 &  \textbf{41.67$\pm$0.7} \\
Ionosphere & 68.12$\pm$0.42& 66.52$\pm$1.1  & 67.04$\pm$0.79 & 71.23$\pm$0.91  & 70.9$\pm$0.99 & 70.52$\pm$0.15   &  \textbf{73.67$\pm$0.41}\\
Iris  & 73.51$\pm$0.82 & 79.92$\pm$1.86  & 80.39$\pm$0.83 & 85.62$\pm$0.82  & 75.31$\pm$0.28& 92$\pm$0.59   &  \textbf{96.3$\pm$0.62}\\
Letters  & 42.82$\pm$0.81& 48.95$\pm$1.34  & 47.68$\pm$0.98 & 54.32$\pm$0.9& 52.19$\pm$0.49  & 53.69$\pm$0.73&  \textbf{55.83$\pm$0.26} \\
MNIST  & 52.18$\pm$2.76& 55.66$\pm$1.41  & 62.46$\pm$0.76 & 59.92$\pm$1.41& 67.39$\pm$0.97  & 66.21$\pm$0.92&   \textbf{69.72$\pm$0.71} \\
Optdigit & 65.92$\pm$1.2& 70.27$\pm$0.84  & 74.67$\pm$0.42 & 78.99$\pm$1.02  & 76.69$\pm$0.72& 77.16$\pm$0.21      &  \textbf{80.56$\pm$0.69} \\
Pendigits  & 52.88$\pm$0.92& 55.73$\pm$0.75  & 54.08$\pm$0.38 & 62.82$\pm$0.81  & 60.78$\pm$0.95 & 61.68$\pm$0.18      &  \textbf{64.13$\pm$0.42}\\
Reuters-21578 & 62.34$\pm$0.72 & 70.24$\pm$0.92  & 71.82$\pm$0.78& 74.63$\pm$0.87  & 75.29$\pm$0.66 & 68.85$\pm$0.32 &  \textbf{76.41$\pm$0.24} \\										
SA Hart  & 66.39$\pm$1.62& 67.38$\pm$1.02  & 66.53$\pm$1.26 & 70.54$\pm$0.93  & 71.42$\pm$0.87& \textbf{73.7$\pm$0.46} &  72.05$\pm$0.16\\
Sonar  & 50.48$\pm$0.92& 53.84$\pm$1.01  & 53.25$\pm$0.51 & \textbf{61.82$\pm$0.72}  & 59.12$\pm$0.83 & 54.39$\pm$0.25  &  60.06$\pm$0.87\\
Statlog  & 52.28$\pm$0.91& 55.39$\pm$0.75  & 55.26$\pm$0.97 & 57.33$\pm$0.91  & 56.42$\pm$0.92& 55.77$\pm$0.71  &  \textbf{59.76$\pm$0.5}\\
USPS   & 60.49$\pm$0.84 & 59.42$\pm$0.78  & 62.11$\pm$0.37 & 64.92$\pm$1.68  & 63.08$\pm$0.59 & 65.21$\pm$0.69   &  \textbf{66.01$\pm$0.24}\\					
Wine & 70.24$\pm$0.72 & 75.62$\pm$1.79  & 81.25$\pm$0.93 & 79.29$\pm$0.51  & 83.16$\pm$0.84& 71.34$\pm$0.55 &  \textbf{89.46$\pm$0.14} \\
Yeast  & 33.81$\pm$0.32 & 36.23$\pm$0.61  & 35.23$\pm$0.72 & 40.25$\pm$0.88  & \textbf{42.03$\pm$0.91}& 37.76$\pm$0.26 &   41.12$\pm$0.4\\
\hline
\end{tabular}\end{sc}\end{small}
\end{center}
\vskip -0.1in
\end{table*}
\begin{figure*}[ht]
	\vskip 0.2in
	\begin{center}
		\begin{minipage}{0.24\linewidth}
			\includegraphics[width=0.95\textwidth]{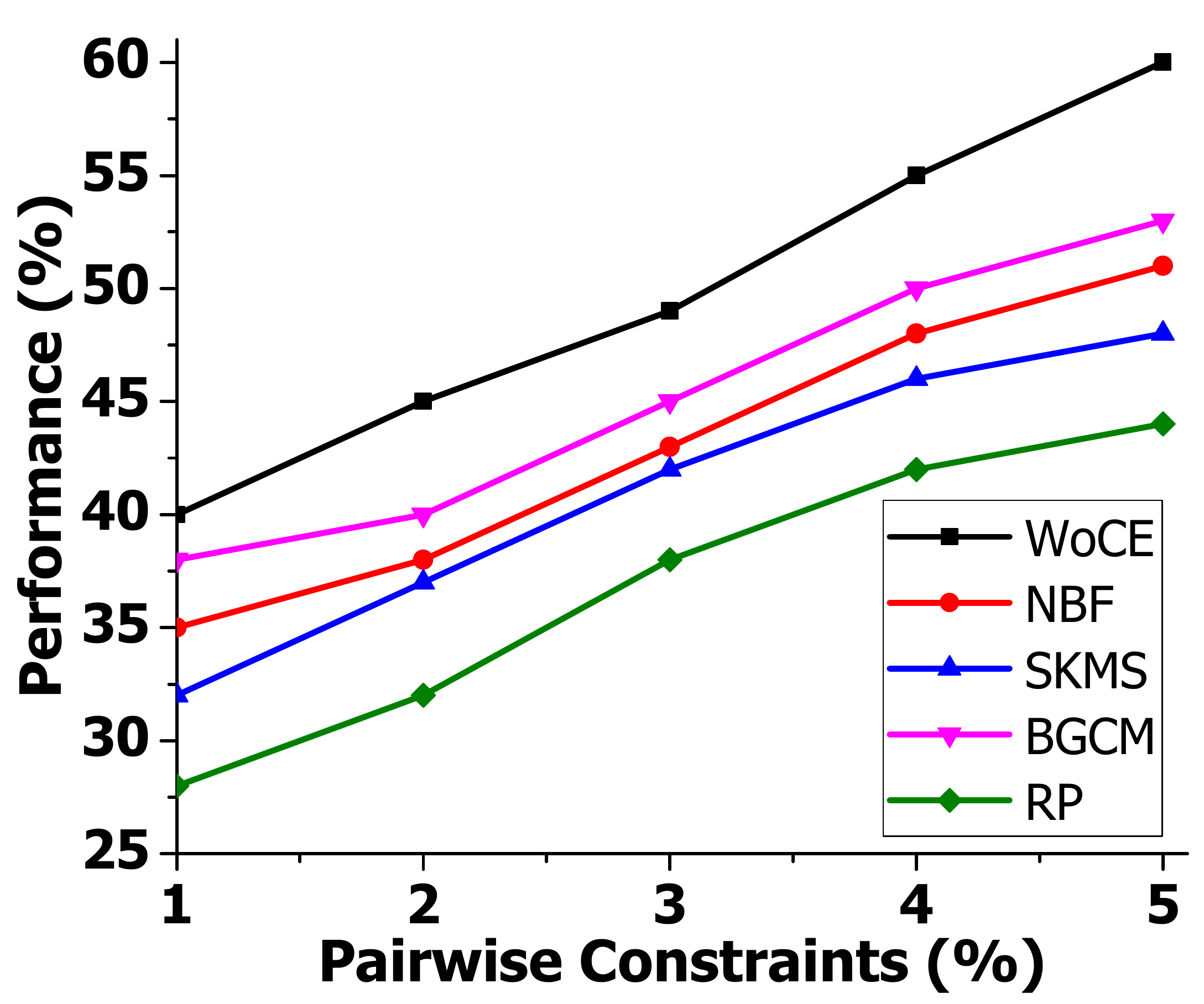}\\
			\centering (a) 20 Newsgroups
		\end{minipage}
		\begin{minipage}{0.24\linewidth}
			\includegraphics[width=0.95\textwidth]{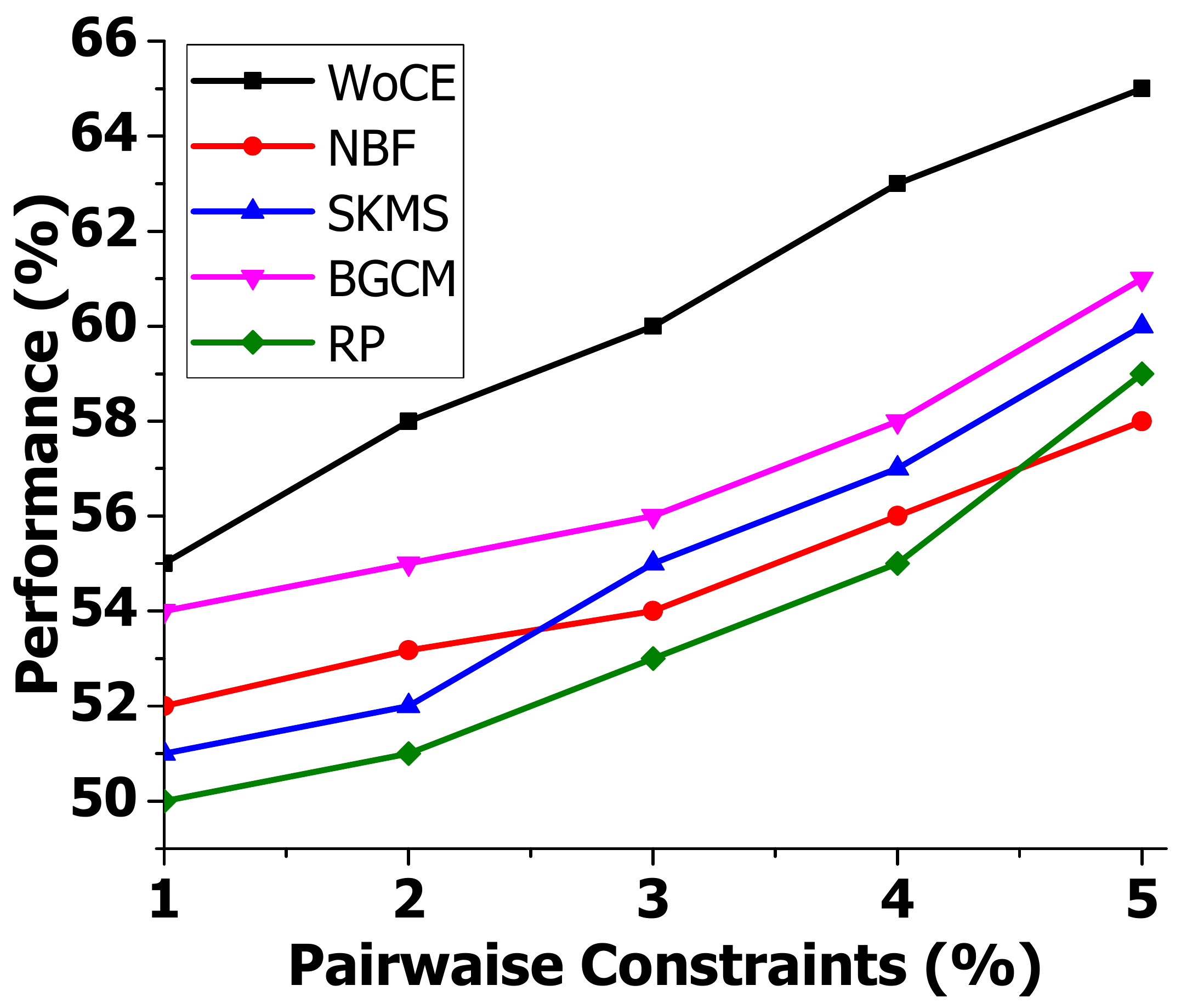}\\
			\centering (b) ADNI-FUL-C1
		\end{minipage}
		\begin{minipage}{0.24\linewidth}
			\includegraphics[width=0.95\textwidth]{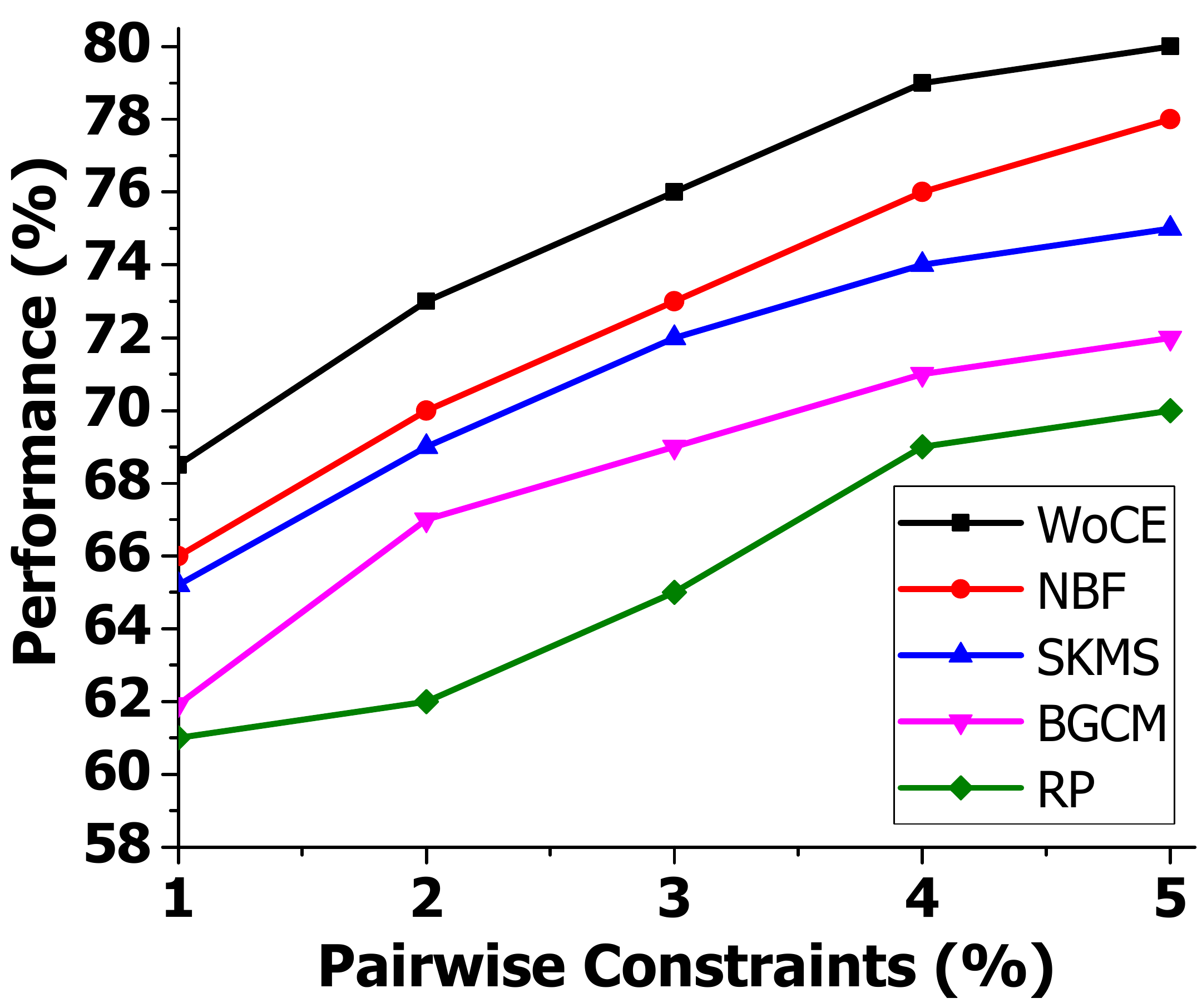}\\
			\centering (c) Arcene
		\end{minipage}
		\begin{minipage}{0.24\linewidth}
			\includegraphics[width=0.95\textwidth]{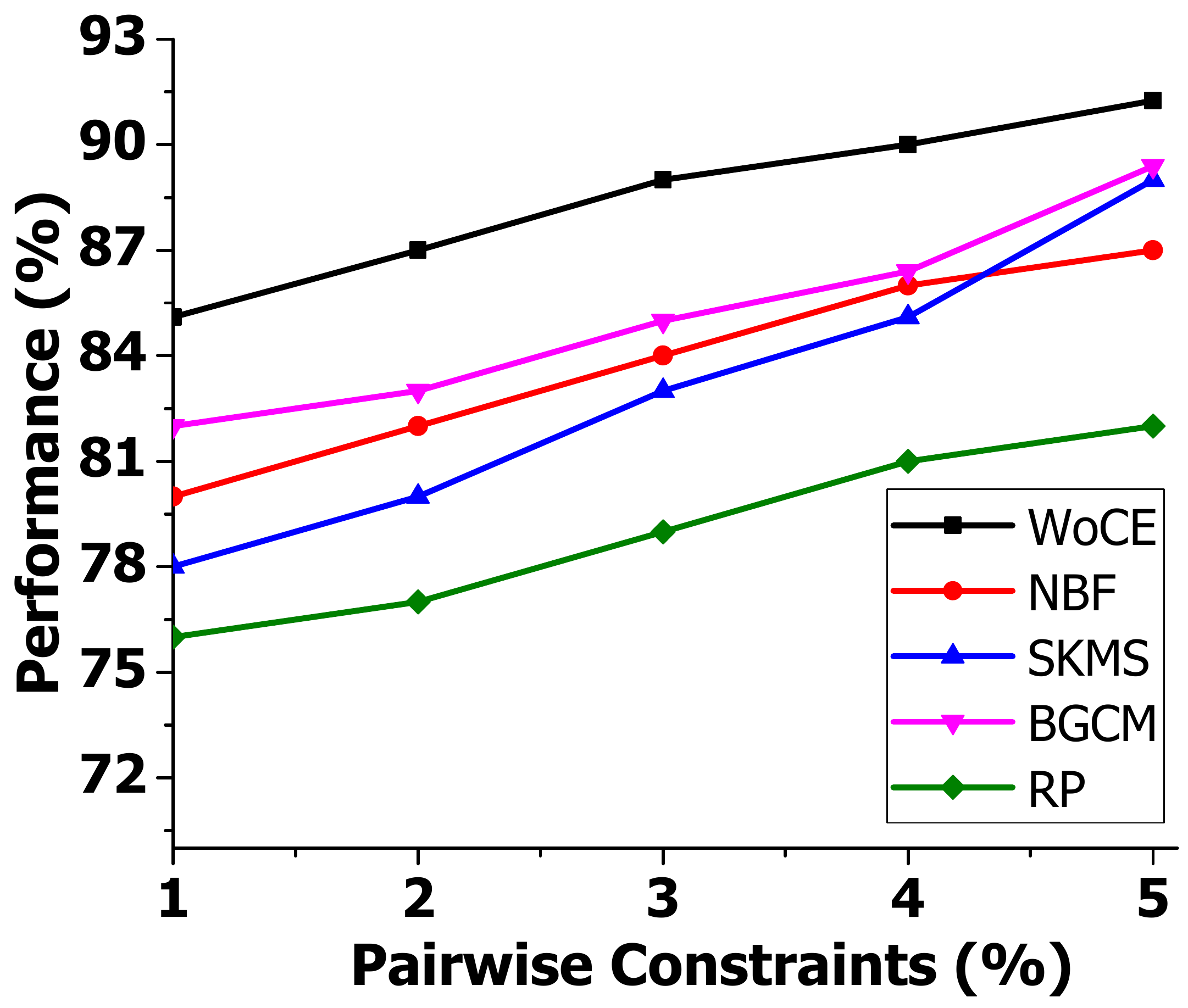}\\
			\centering (d) CNAE-9 
		\end{minipage}
		\begin{minipage}{0.24\linewidth}
			\includegraphics[width=0.95\textwidth]{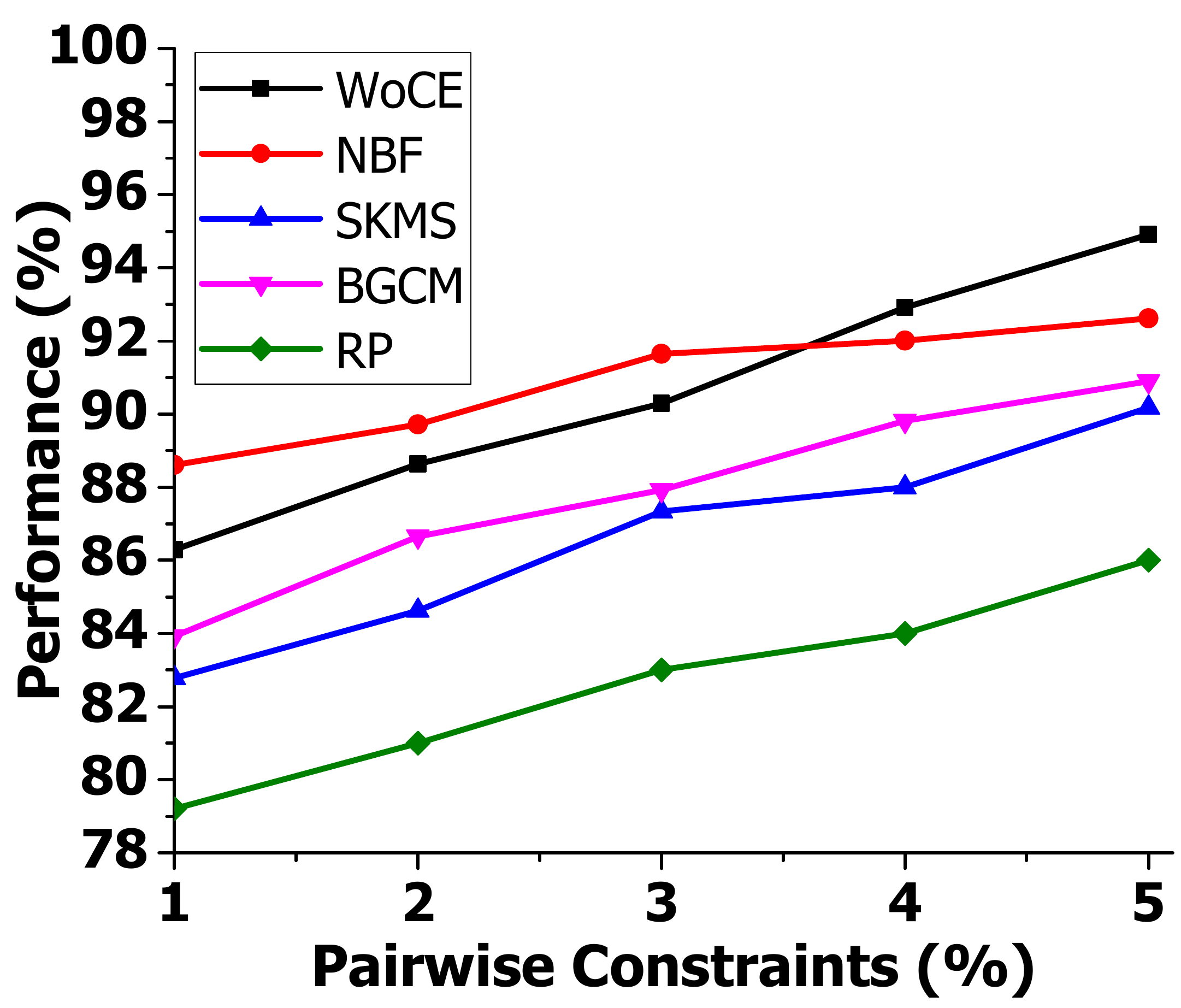}\\
			\centering (e) Optdigit
		\end{minipage}
		\begin{minipage}{0.24\linewidth}
			\includegraphics[width=0.95\textwidth]{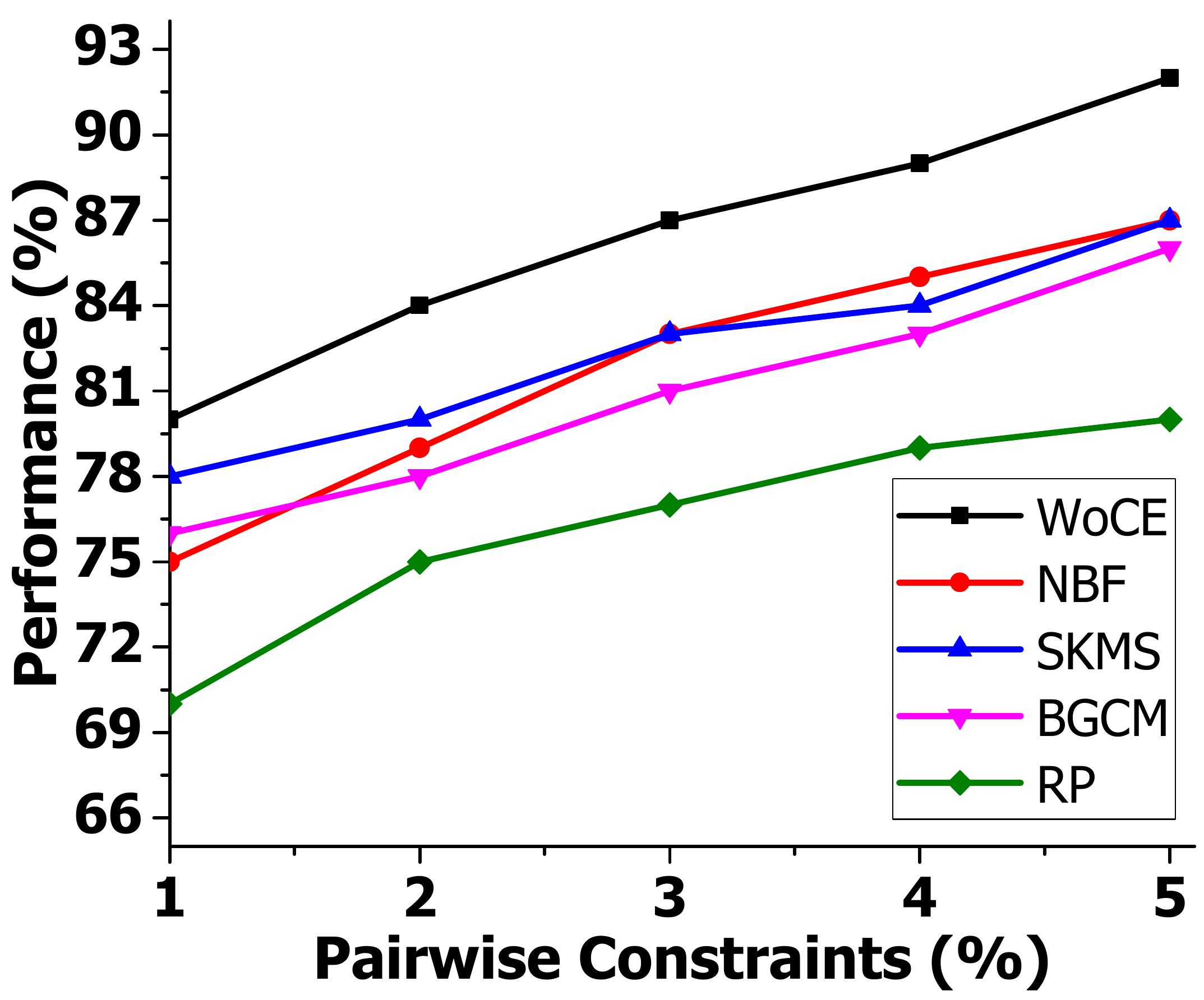}\\
			\centering (f) Reuters-21578
		\end{minipage}
		\begin{minipage}{0.24\linewidth}
			\includegraphics[width=0.95\textwidth]{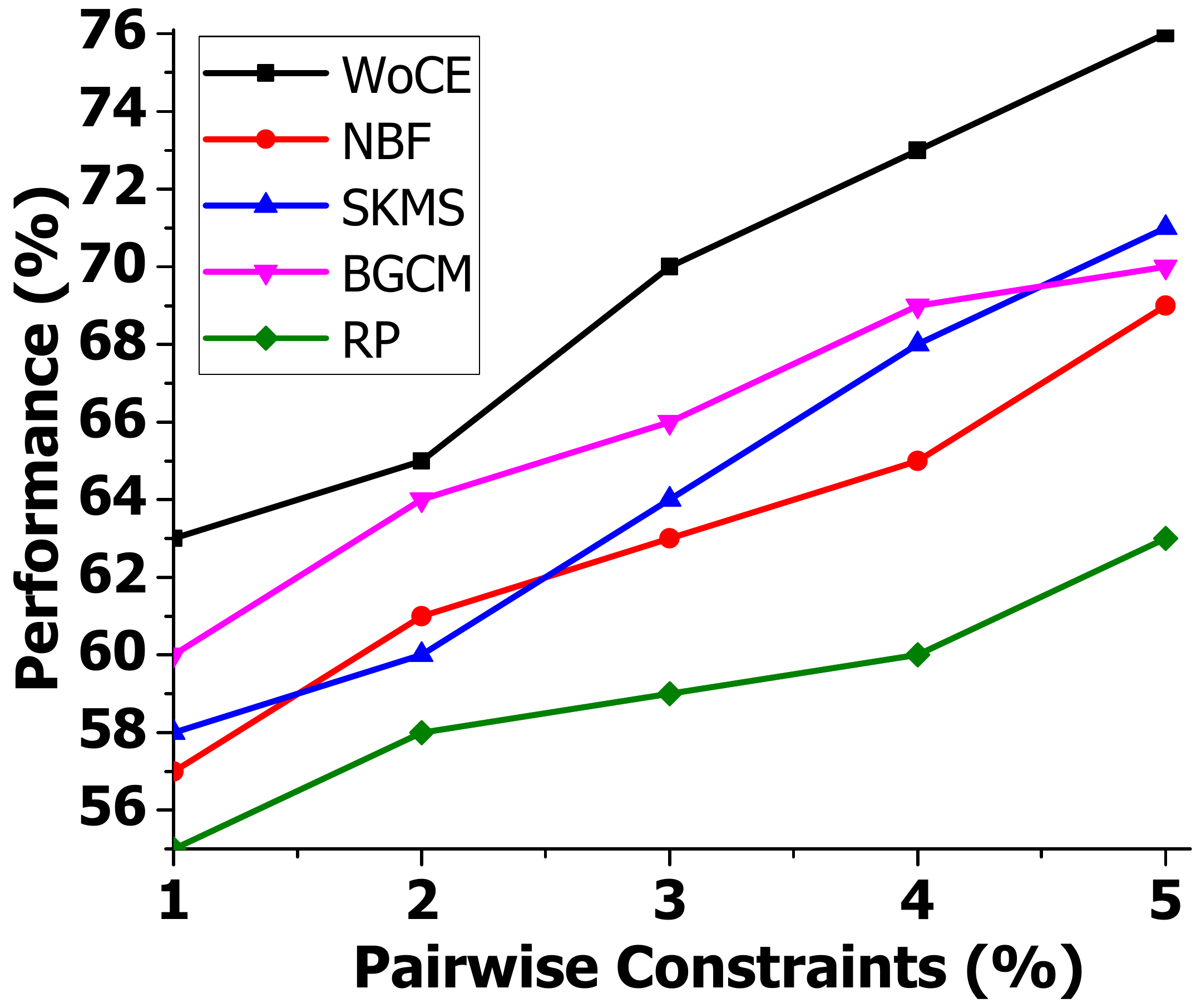}\\
			\centering (g) Sonar
		\end{minipage}
		\begin{minipage}{0.24\linewidth}
			\includegraphics[width=0.95\textwidth]{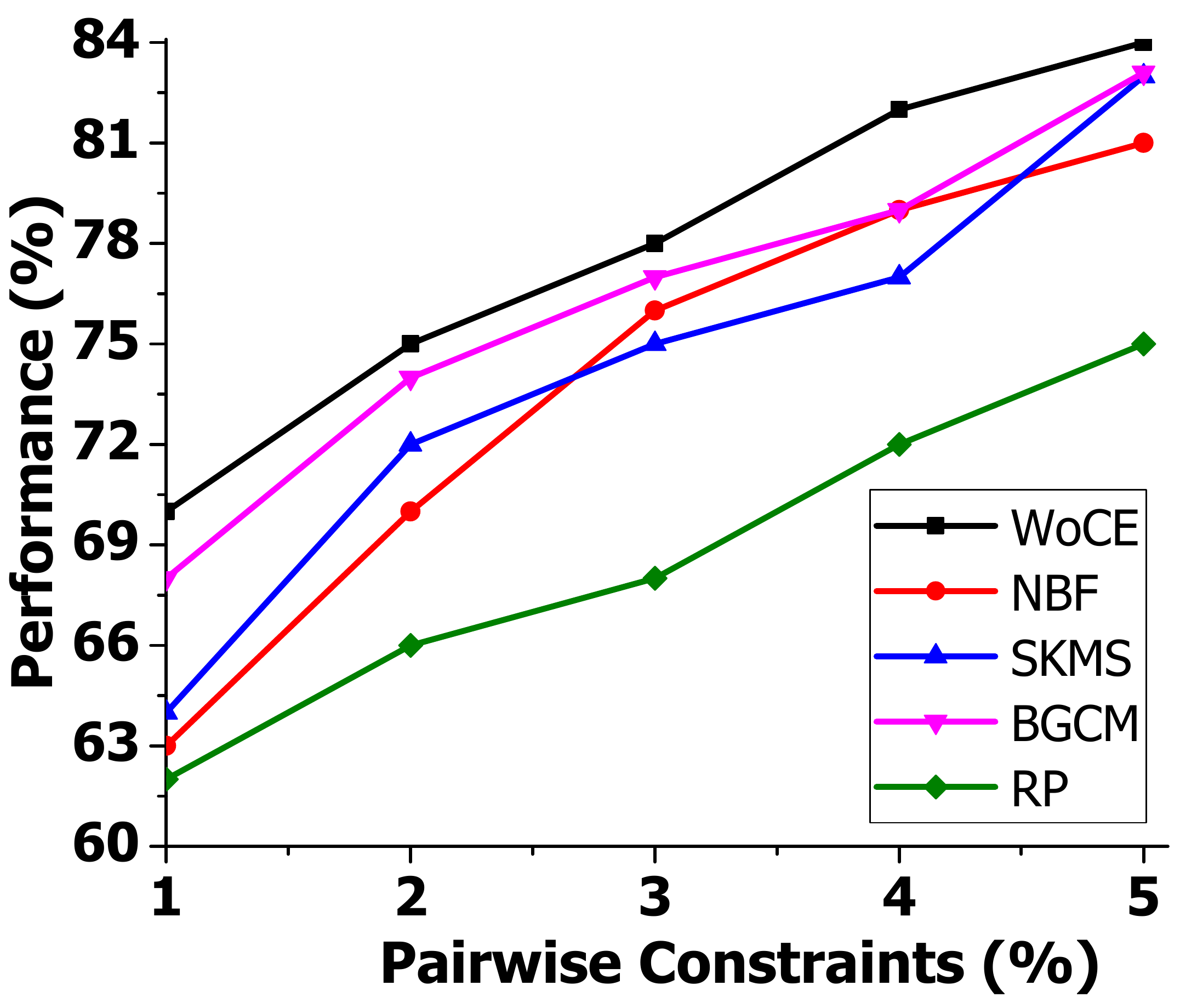}\\
			\centering (h) USPS
		\end{minipage}
		\begin{minipage}{0.24\linewidth}
	\includegraphics[width=0.95\textwidth]{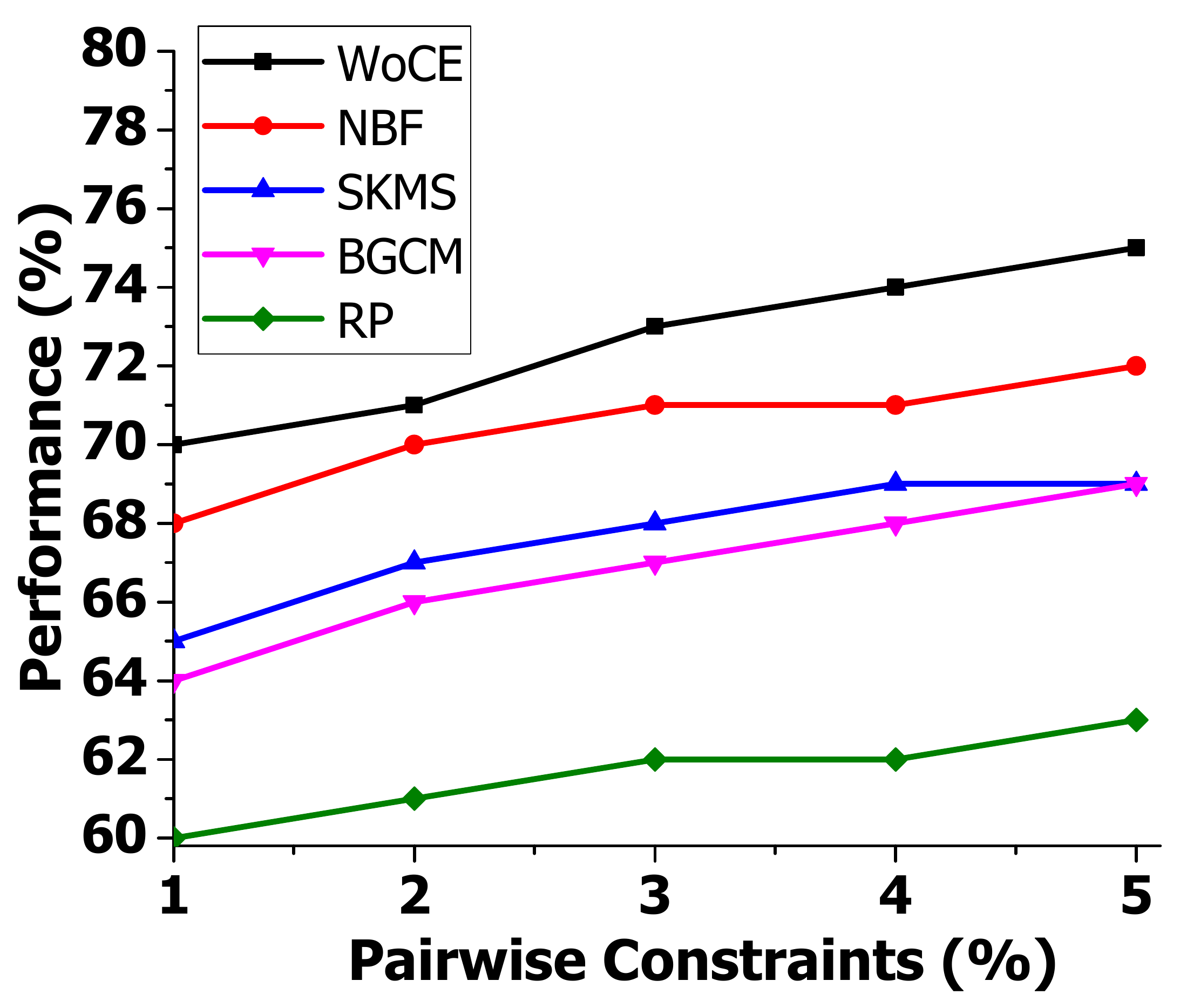}\\
	\centering (i) CIFAR-10
\end{minipage}	
		\begin{minipage}{0.24\linewidth}
	\includegraphics[width=0.95\textwidth]{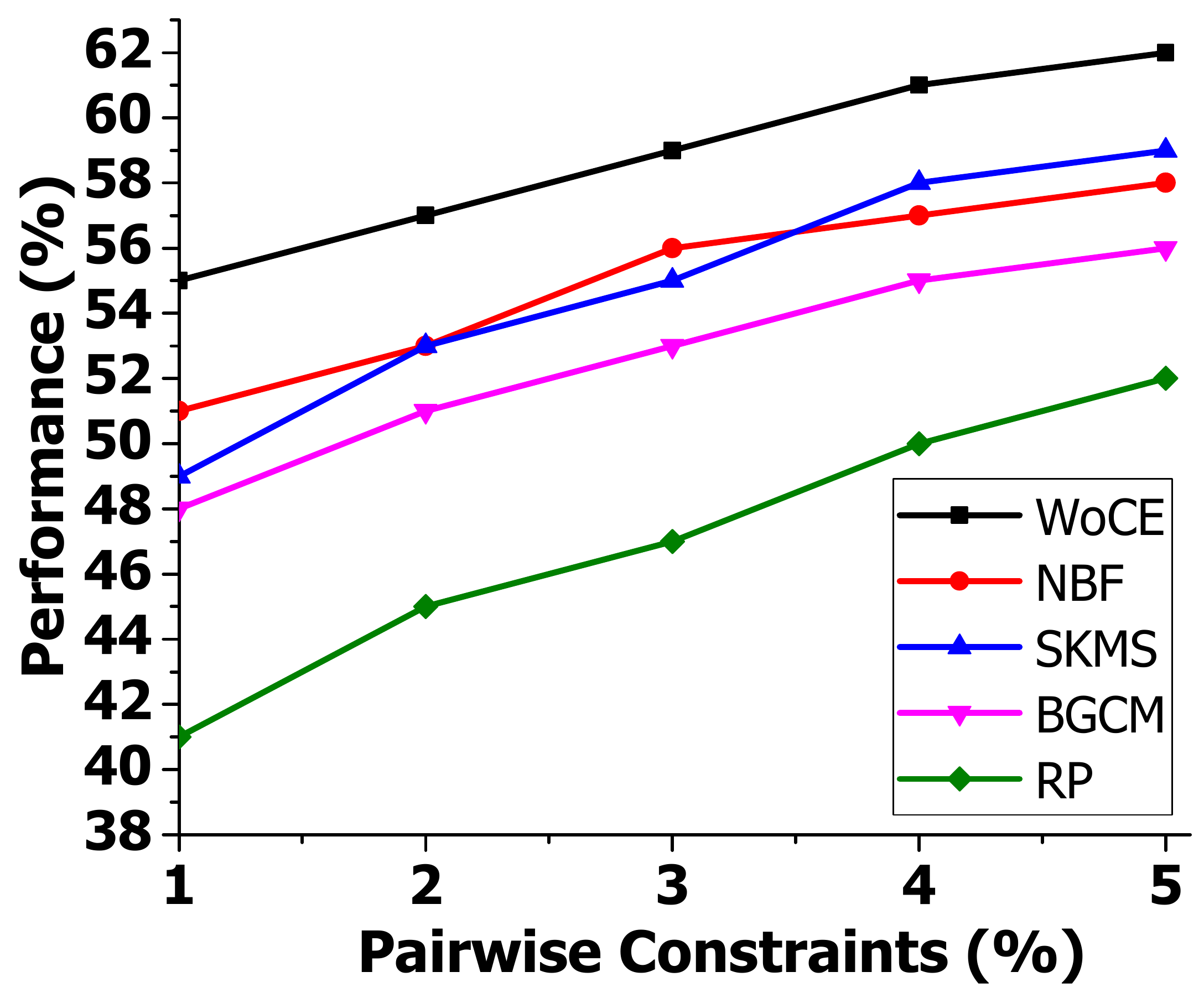}\\
	\centering (j) ImageNet
\end{minipage}	
		\begin{minipage}{0.24\linewidth}
	\includegraphics[width=0.95\textwidth]{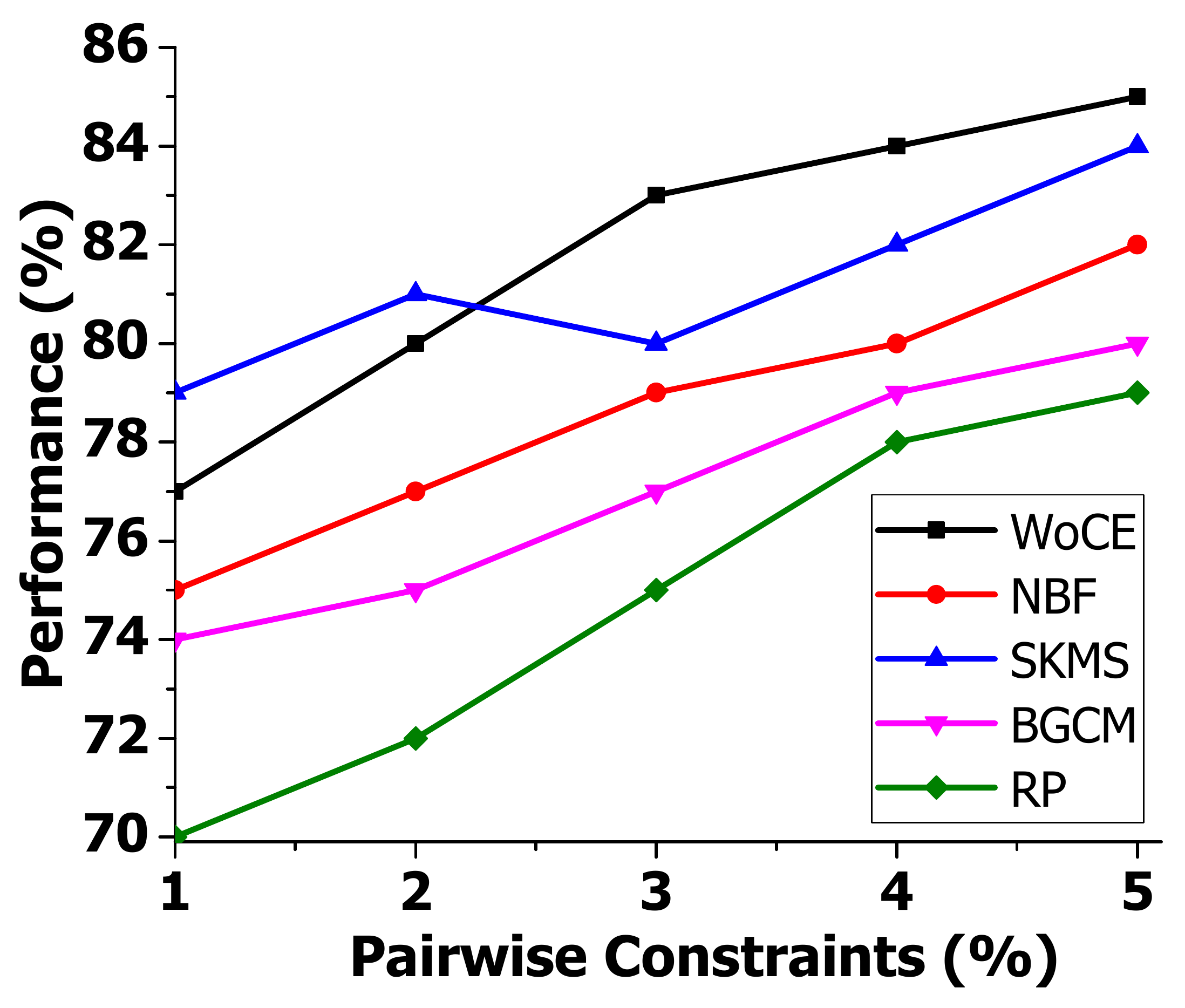}\\
	\centering (k) MNIST
\end{minipage}	
		\begin{minipage}{0.24\linewidth}
	\includegraphics[width=0.95\textwidth]{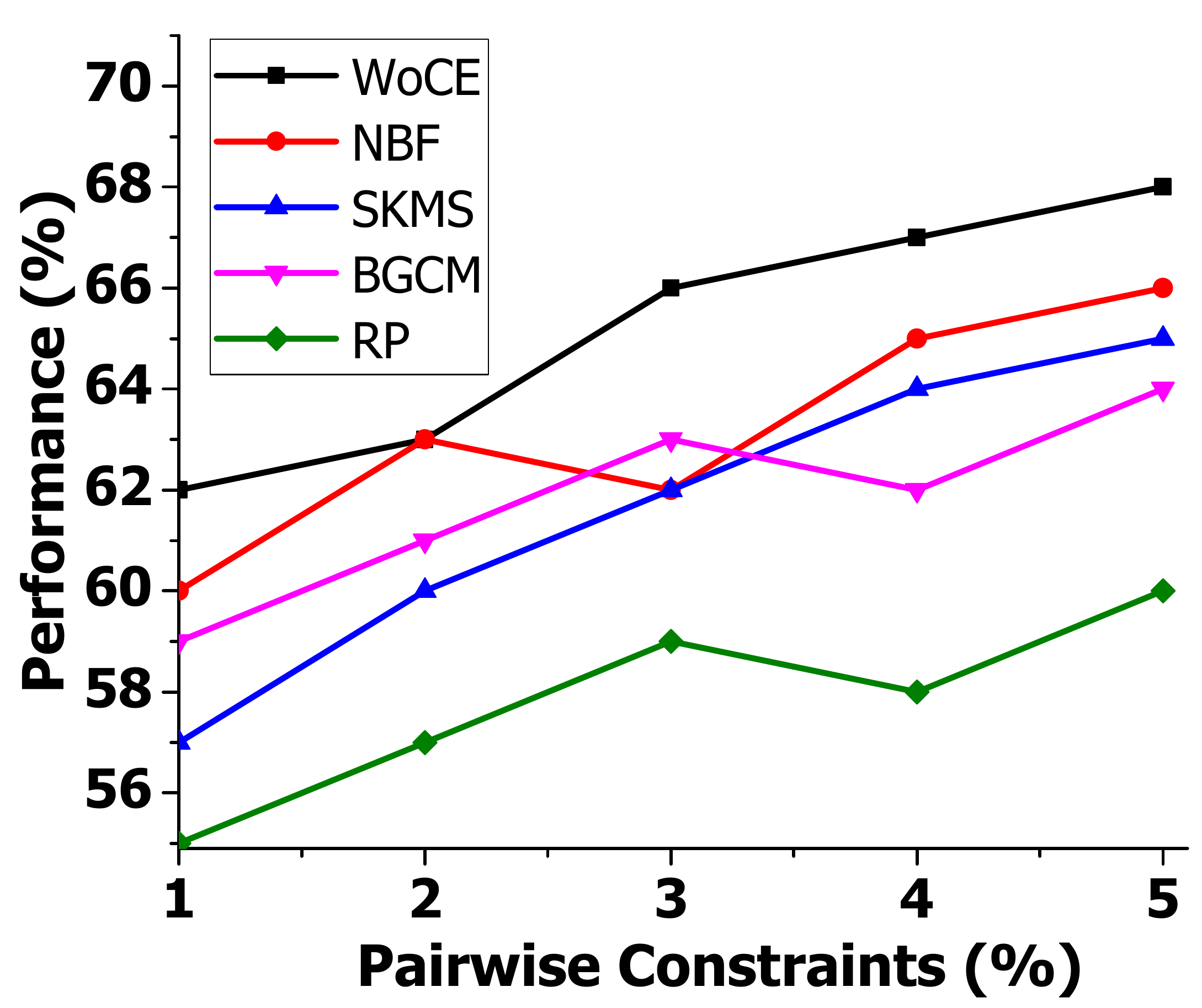}\\
	\centering ($\ell$) Letters
\end{minipage}				
		\caption{The performance of semi-supervised methods.}
		\label{Fig2}
	\end{center}
	\vskip -0.2in
\end{figure*}
\subsection{Performance analysis for semi-supervised methods}
The empirical results of semi-supervised methods will be analyzed in this section. Since most of the semi-supervised cluster ensemble methods \cite{Xiong14,Liu14,Gao13} use feature selection based on the supervision information, this paper compares the performance of semi-supervised methods on high-dimensional and large-scale data set in the Table II; i.e. 20 Newsgroups, Letters, and Reuters-21578 as document-based data sets, ADNI, CIFAR-10, ImageNet, MNIST, and USPS as image-based data sets, and also Arcene, CNAE-9, Optdigit, Sonar from UCI repository. This paper does not use the optional feature selection in this section ($d = 0$). 

In this paper, 1\% to 5\% of instances with class labels are randomly selected for generating the supervision information (a half for must-link and a half for cannot-link); e.g. 1\% (2620) of instances are selected in the 20 Newsgroups data set for generating the pairwise constraints, where 655 must-links and 655 cannot-links constraints are generated by the selected instances. In addition, the supervision information, which applied to the methods, are same in each independent run for all methods. Notably, this paper does not employ all combinations of the randomly selected instances as the pairwise constraints (must-links and cannot-links). In other words, each randomly selected instance is used once for generating just a must-link or a cannot-link. There are two reasons for this strategy of generating pairwise constraints. Firstly, this strategy provides better diversity among the generated pairwise constraints. Secondly, this strategy represents better simulation for the real application of prior supervision information. Indeed, there is no class label in the real-world applications, and generating the pairwise constraints from all combinations of the randomly selected instances is impossible or expensive \cite{Anand14,Xiong14,Gao13}. For instance, just consider an interactive image search engine, where it shows two random images to users in each attempt and asks users to specify that these images are same (must-link) or different (cannot-link). Then, the search engine will improve the clustering results based on these limited feedbacks.  

The final clustering performance (accuracy metric \cite{Tan05}) was evaluated by re-labeling between obtained clusters and the ground truth labels and then counting the percentage of correctly classified samples \cite{Alizadeh15}. Figure \ref{Fig2} illustrates the performance of the proposed method (WoCE) in comparison with the RP \cite{Ho14}, BGCM \cite{Gao13}, NBF \cite{Xiong14} and SKMS \cite{Anand14}. In this figure, the standard deviations of the results are lower than 1\% (for 10 independent runs). This paper reports the performance of RP as a classical method in the semi-supervised cluster ensemble. Also, this paper reports the performance of BGCM as a novel graph-based approach in the semi-supervised clustering. Notably, the BGCM has two versions; i.e. unsupervised and semi-supervised. This paper uses the semi-supervised version of BGCM in this section. What's more, this paper uses SKMS as a kernel-based method in semi-supervised clustering. Last but not least, the empirical results of the proposed method are compared with NBF as another heuristic method in the semi-supervised cluster ensemble. Even though WoCE was outperformed in one data set (Optdigit) by some algorithms, the majority of results demonstrate superior accuracy for the proposed method. In addition, the clustering performance of some algorithms in Fig. \ref{Fig2} (k) and ($\ell$) become worse with increased number of pairwise constraints. As mentioned before, pairwise constraints often result in highly unstable clustering performance \cite{Anand14,Xiong14}. These figures are good examples for this issue, where some of the previous methods cannot handle the extra supervision information. In fact, the supervision information made unstable individual clustering results and significantly reduce the performance of the mentioned methods. In these cases, our proposed method has handled the supervision information by employing the WOC theory, i.e. better data representation (Algorithm 1 and 2), robust individual clustering evaluation (Uniformity metric), and effective aggregating mechanism (WEAC).
\subsection{Optional feature selection in unsupervised-method}
\begin{figure}[ht]
	\vskip 0.2in
	\begin{center}
		\begin{minipage}{0.48\linewidth}
			\includegraphics[width=\textwidth]{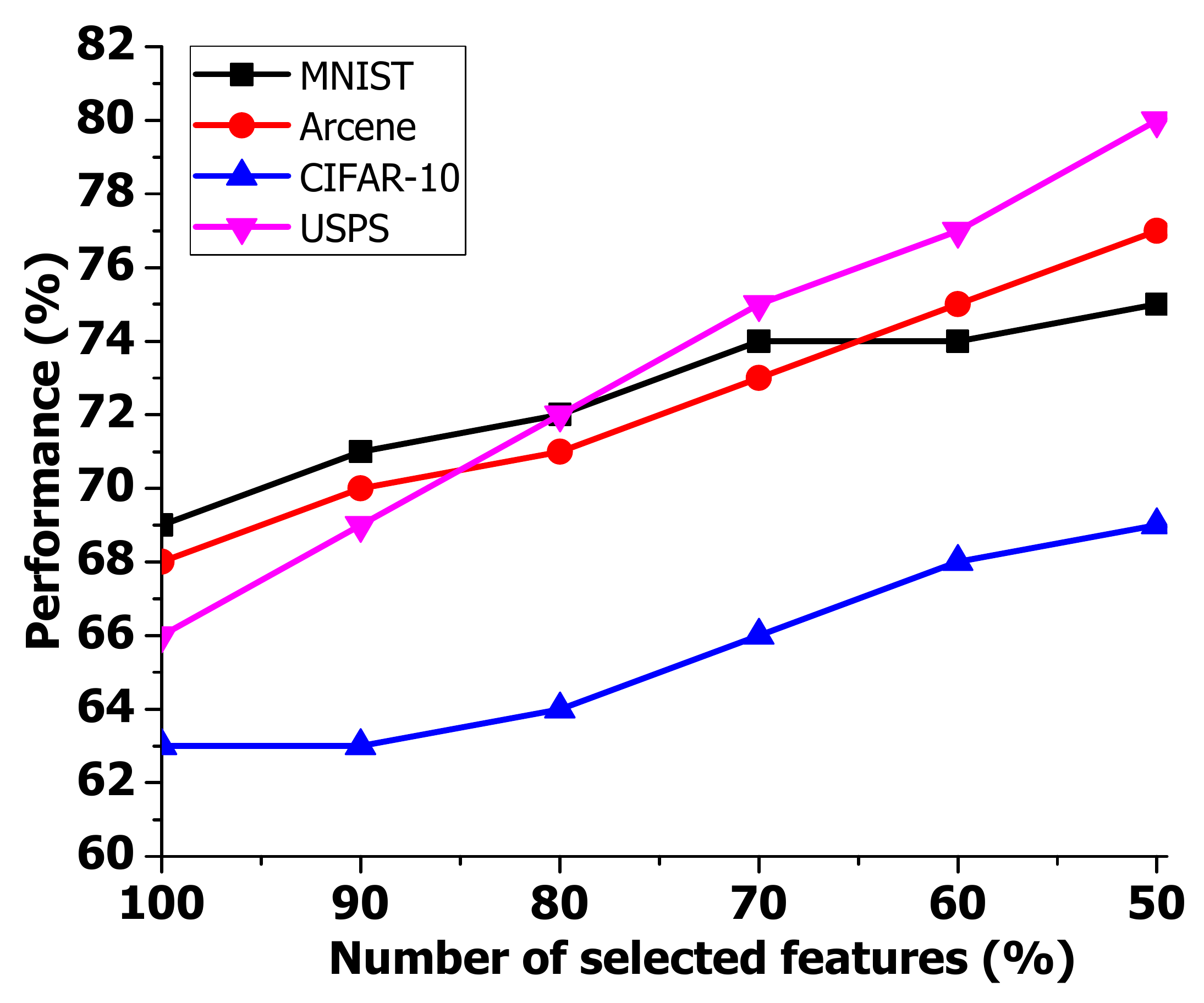}\\
			\centering (a)
		\end{minipage}
		\begin{minipage}{0.48\linewidth}
			\includegraphics[width=\textwidth]{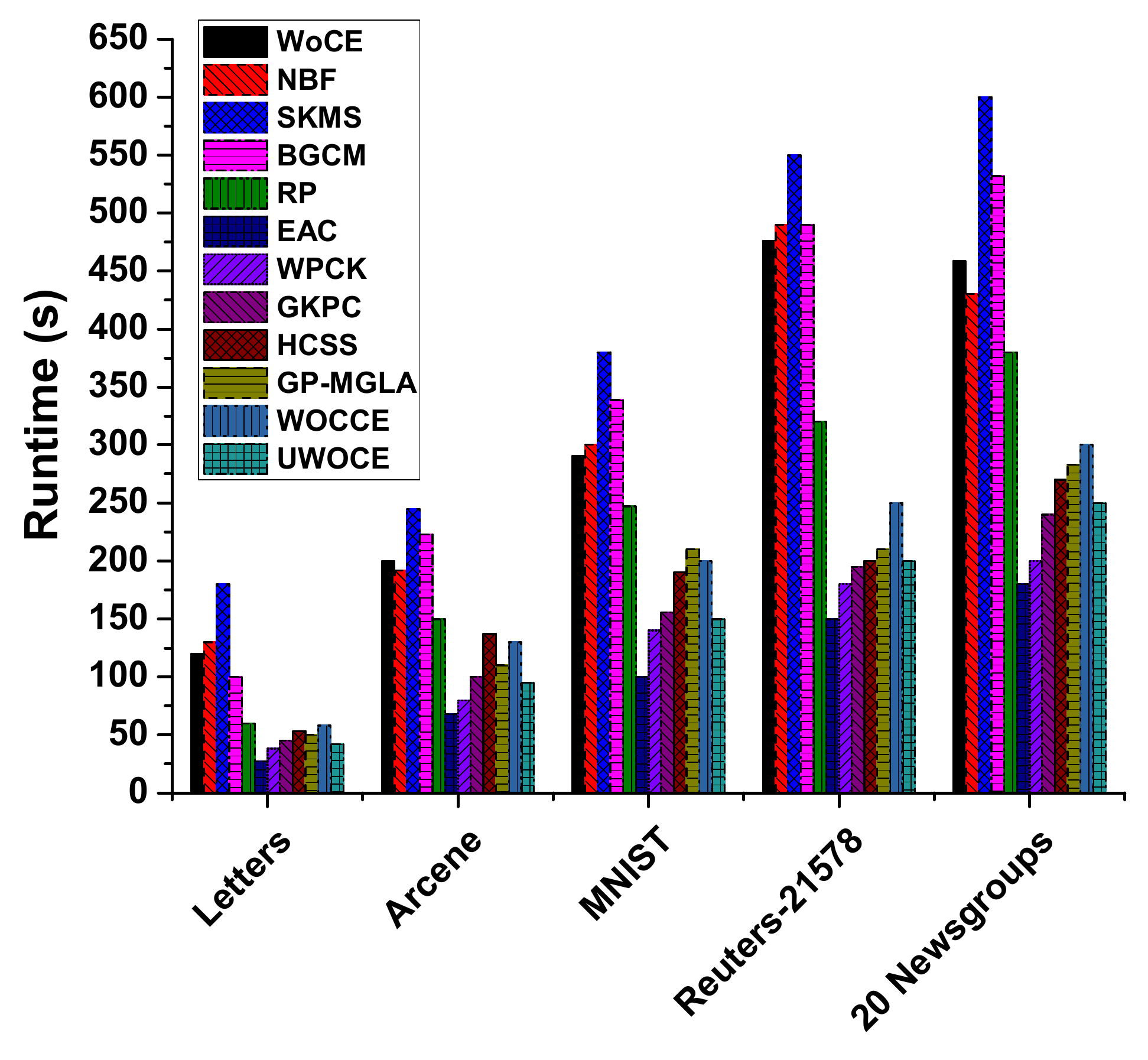}\\
			\centering (b)
		\end{minipage}
			\caption{(a) The performance of UWoCE method by using the optional feature selection. (b) The runtime analysis.}
	\label{Fig3}
	\end{center}
\vskip -0.2in
\end{figure}

In this section, the performance of the proposed method will be analyzed by using the optional feature selection ($d$ parameter). Since feature selection is automatically used by applying the supervision information on the mapped data set in the semi-supervised version of the proposed method, the performance of the unsupervised version of the proposed method (UWoCE) will be analyzed in this section. This paper employs high-dimension data sets, i.e. Arcene, CIFAR-10, MNIST, and USPS, for analyzing the performance of the proposed method. Figure \ref{Fig3} (a) shows the relationship between the performances of the proposed method based on the percentage of selected features in different data sets. The vertical axis refers to the performance while the horizontal axis refers to the percentage of the selected features in each data set. As depicted in this figure, the optional feature selection can significantly increase the performance of final results on high-dimensional data sets. So, this paper offers that the optional feature selection will be used for high-dimensional data sets to handle features' sparsity. Moreover, this experiment is the reason for using high-dimensional and large-scale data sets in the previous section. The important questions which must be discussed here are what is different between the mapping function and the optional feature selection? And where the optional feature selection can be employed? Indeed, the mapping function, which is illustrated in Algorithm 1, minimizes the correlation between features; and it maps data to a new domain, which the covariance between different features is near zero. Most of the time, this function maps data to the stable dimensions, which can dramatically improve the accuracy of the final results. It can be also formulated as follows: $Q: X\in \mathbb{R}^{m\times n} \to Y\in \mathbb{R}^{m\times n}$, where $Q$ can satisfy the independency criterion in the WOC theory. In a high-dimensional data set, some of the calculated eigenvalues ($\Lambda = \{\lambda_j \}$) approach near to zero. Since these eigenvalues trivially effect on the mapping $Q$, they can be omitted for reducing the dimensions of the data set and the time and space complexities of the clustering analysis. In other words, these eigenvalues may reduce the stability of the final results \cite{Yousefnezhad15}. By considering the optional feature selection, the mapping $Q$ can be formulated as follows: $Q: X\in \mathbb{R}^{m\times n} \to Y\in \mathbb{R}^{d\times n}$, where $d < m$. Therefore, employing this optional feature selection for analyzing the high-dimension data set can improve the stability of the mapping $Q$ as well as the performance of the final result (see Fig. \ref{Fig3} (a) ). In addition, this feature selection is better to use based on the fluctuation of the eigenvalues (remove near zero values) in each problem.
%\begin{figure}[!t]
%	\centering
%	\includegraphics[width=0.23\textwidth]{Images/Fig3}
%    \caption{The performance of UWoCE method by using the optional feature selection.}
%    \label{Fig3}
%    \vskip -0.3in	
%\end{figure}
%\begin{figure}[!h]
%	\includegraphics[width=0.24\textwidth]{Images/Fig4}
%	\centering
%	\caption{The runtime analysis.}
%     \label{Fig4}
%   \vskip -0.2in	
%\end{figure}	
%\subsection{Noise and Missed-values Analysis}
\subsection{Time complexity analysis}
In this section, the runtime of both unsupervised and semi-supervised methods will be compared by using various data sets, i.e. three large-scale data sets (Letters, MNIST, 20 Newsgroups) and two high-dimension data sets (Arcene, Reuters-21578). Figure \ref{Fig3} (b) illustrates the relationship between the runtime of the mentioned methods and the size of data sets. The vertical axis refers to the runtime while the horizontal axis refers to the algorithms. As mentioned before, all of the results in this experiment are generated by a PC with a certain specifications. As depicted in this figure, the runtime of the semi-supervised methods (the first five bars) is more than the runtime of the unsupervised methods because they need an additional step to apply the supervision information (mostly in the form of an Eigenproblem) \cite{Gao13}. By considering the performance of these methods in Table \ref{Tbl3: Unsupervised Expermintal Results} and Fig. \ref{Fig2}, WoCE (the first bar) and UWoCE (the last bar) generated more efficient results in comparison with other clustering methods. Indeed, the proposed method selects the features based on the correlations between data points and supervision information (in semi-supervised approach). So, the number of calculations for generating individual clustering results will be significantly decreased in comparison with other cluster ensemble methods, while the performance of the final results is rapidly increased. 

There are some technical issues that must be discussed here. Firstly, this paper uses an EM algorithm \cite{tipping99} for estimating the eigenvalues/vectors, which this algorithm can significantly reduce the time complexity of the mapping function in Algorithm \ref{alg:Mapping}. Secondly, the size of the transformed matrix (eq. \refeq{eq:Transformed data}) in the proposed method for applying the supervision information is limited to the size of pairwise constraints. This size is really small in comparison with the size of instances; e.g. In 20 Newsgroup data set, the size of this matrix for 1\% of randomly sampled pairwise constraints is $655\times 655$, while the instance similarity matrix is $26214\times 26214$. Notably, most of the previous studies such as SKMS and BGCM directly used the instance similarity matrix for applying the supervision information. Lastly, this paper uses a modified version of the EAC for combining the individual clustering results. EAC applies a linkage method on a simple matrix, where the size of this matrix is the number of algorithms $\times$ the number of instances ($T\times n$), where $T << n$ in practice. By contrast, some of the previous studies such as BGCM utilized the graph methods for combining the individual results, where the size of the adjacency matrix of the graph in these methods is the square of the number of instances ($n^2$). Based on these technical issues, the proposed method can significantly increase the performance of the final results as well as an acceptable runtime.  
\section{Conclusion}
In this paper, wisdom of crowds (WOC) theory in social science was mapped to the clustering ensemble arena. The main advantages of this mapping include the addition of two new aspects, i.e., independency and decentralization, for estimating the quality of individual clustering results, and a new framework to investigate them. To reach the four conditions of WOC, this paper incorporates a series of novel strategies for producing individual clustering results as well as obtaining the final ensemble result. Specifically, a mapping function is introduced to perform \emph{independency} on individual clustering results. This function can minimize the correlation between features by using the concepts of expected value and covariance. The decentralization criterion is proposed for transforming the data from high-dimension to low-dimension based on pairwise constraints, to keep quality in the generated individual clustering results. Further, this paper evaluates the \emph{diversity} of individual clustering results with a novel metric called uniformity. At last, weighted EAC is proposed for the final \emph{aggregation}. To validate the effectiveness of the proposed approach, an extensive experimental study is performed by comparing with multiple state-of-the-art methods on various data sets. In the future, we will develop a new version of uniformity based on the concept of expected value instead of using the APMM.
\vskip -0.5in
\section*{Acknowledgment}
We thank the anonymous reviewers for comments. This work was supported in part by the National Natural Science Foundation of China (61422204, 61473149, and 61503182), Jiangsu Natural Science Foundation (BK20130034 and BK2015042628), and NUAA Fundamental Research Funds (NE2013105).
\bibliographystyle{IEEEtran}
\bibliography{WoCE_Edited}
\vskip -0.5in
\begin{IEEEbiography}[{\includegraphics[width=1in,height=1.25in,clip,keepaspectratio]{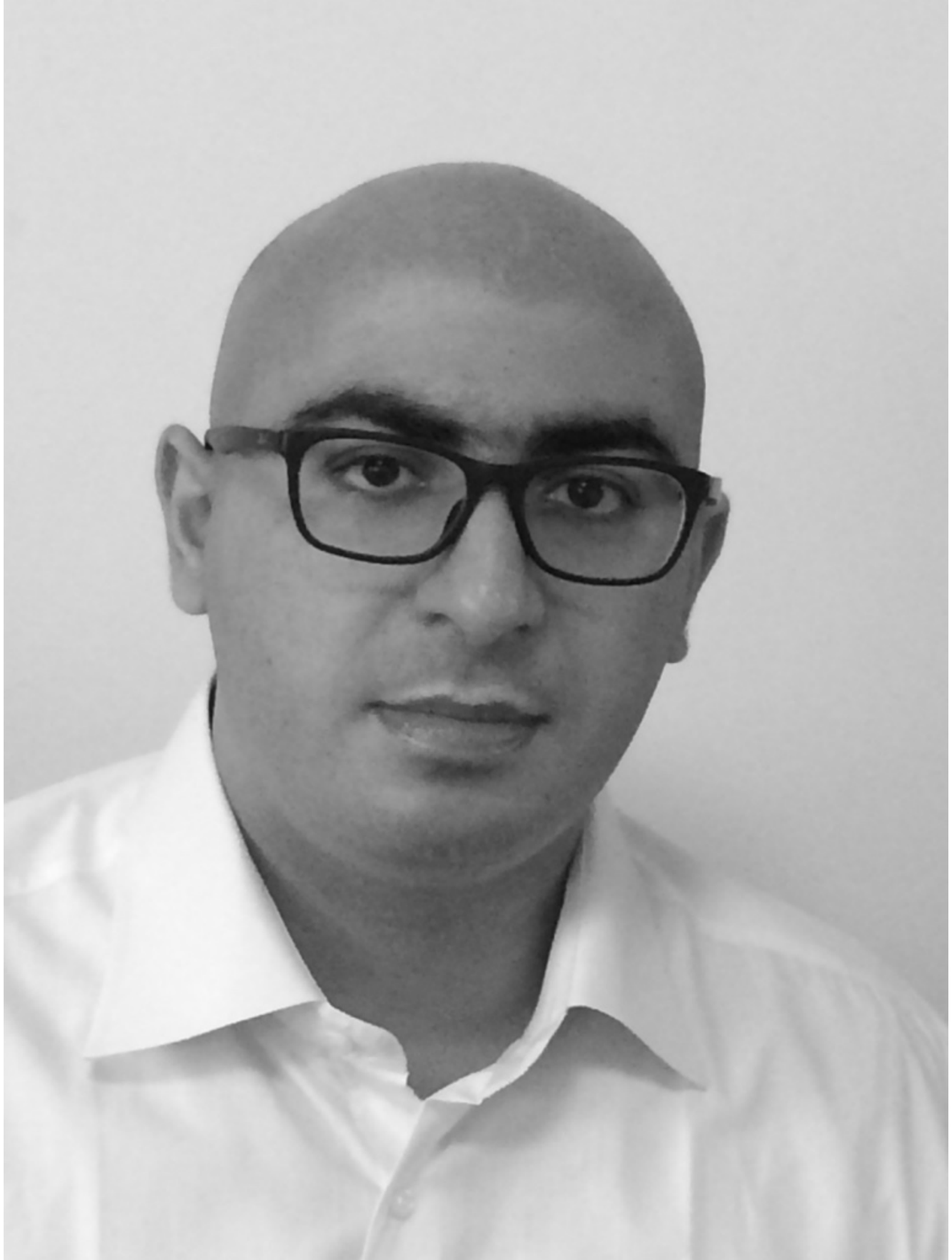}}]{Muhammad~Yousefnezhad} received his B.Sc. and M.Sc. degrees in Computer Hardware Engineering and Information Technology (IT), with a sheer focus on Artificial Intelligence, from Mazandaran University of Science and Technology (MUST), Iran, in 2010 and 2013, respectively. He joined the College of Computer Science and Technology at Nanjing University of Aeronautics and Astronautics as a Research Assistant for his Ph.D. research in 2014. His main research interest is developing machine learning techniques, particularly within the area of the human brain decoding.\end{IEEEbiography}
\vskip -0.5in
\begin{IEEEbiography}[{\includegraphics[width=1in,height=1.25in,clip,keepaspectratio]{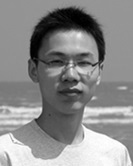}}]{Sheng-Jun~Huang}
received the B.Sc. and Ph.D. degrees in computer science from Nanjing University, China, in 2008 and 2014, respectively. He is currently an Associate Professor in the College of Computer Science and Technology at Nanjing University of Aeronautics and Astronautics. His main research interests include machine learning and patter recognition. He has won the China Computer Federation (CCF) Outstanding Doctoral Dissertation Award in 2015, the Best Poster Award at KDD'12, the Best Student Paper Award at CCDM'11, and the Microsoft Fellowship Award in 2011.
\end{IEEEbiography}
\vskip -0.5in
\begin{IEEEbiography}[{\includegraphics[width=1in,height=1.25in,clip,keepaspectratio]{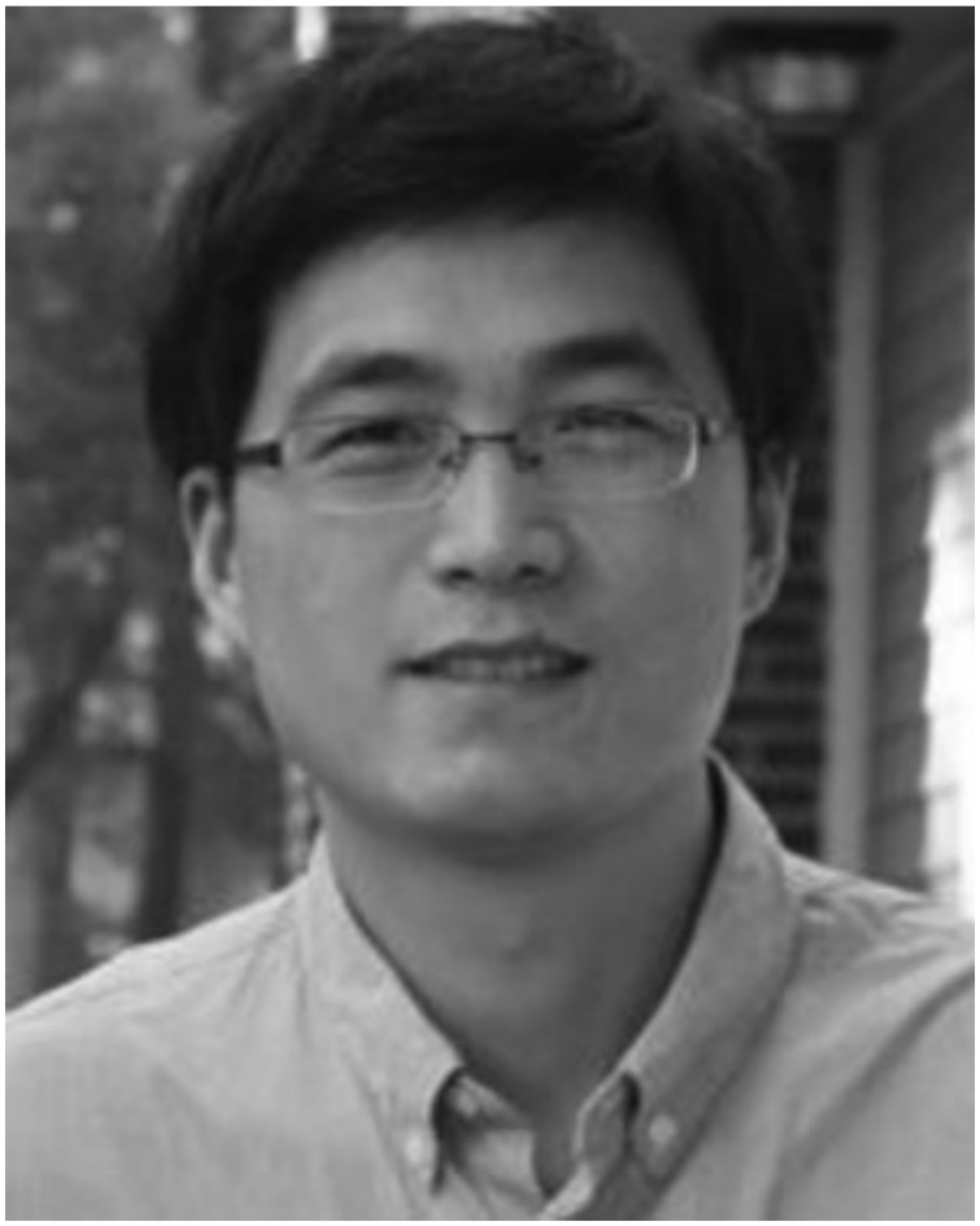}}]{Daoqiang~Zhang}
received the B.Sc. and Ph.D. degrees in computer science from Nanjing University of Aeronautics and Astronautics, Nanjing, China, in 1999 and 2004, respectively. He is currently a Professor in the Department of Computer Science and Engineering, Nanjing University of Aeronautics and Astronautics. His current research interests include machine learning, pattern recognition, and biomedical image analysis. In these areas, he has authored or coauthored more than 100 technical papers in the refereed international journals and conference proceedings. 
\end{IEEEbiography}
\end{document}